\documentclass{bmcart}                     
\usepackage{graphicx}
\usepackage{booktabs} 
\usepackage{pgfplots}
\usetikzlibrary{patterns}
\usepackage{fancyhdr}
\usepackage{multirow}
\usepackage{comment}
\usepackage{textcomp}
\usepackage{graphicx}
\usepackage{lscape}
\usepackage{rotating}
\usepackage{algorithm}
\usepackage{algpseudocode}
\usepackage[utf8]{inputenc}
\begin{document}
\begin{frontmatter}

\begin{fmbox}
\dochead{Survey}
\title {Confidential Machine Learning on Untrusted Platforms: A Survey}



 \author[
       addressref={aff1},                   
       corref={},
       email={sagar.sharma@hp.com}   
    ]{\inits{SS}\fnm{Sagar} \snm{Sharma}}
\author[
       addressref={aff2},     
       corref={},               
       email={keke.chen@marquette.edu}
    ]{\inits{JRS}\fnm{Keke} \snm{Chen}}

\address[id=aff1]{
  \orgname{HP Inc.},
  \city{Vancouver},
  \cny{USA}
}
\address[id=aff2]{
  \orgname{Department of Computer Science, Marquette University},
  \city{Milwaukee},
  \cny{USA}
}



 \begin{abstractbox}

    \begin{abstract}
With the ever-growing data and the need for developing powerful machine learning models, data owners increasingly depend on various untrusted platforms (e.g., public clouds, edges, and machine learning service providers) for scalable processing or collaborative learning. Thus, sensitive data and models are in danger of unauthorized access, misuse, and privacy compromises. A relatively new body of research confidentially trains machine learning models on protected data to address these concerns. In this survey, we summarize notable studies in this emerging area of research.  With a unified framework, we highlight the critical challenges and innovations in outsourcing machine learning confidentially. We focus on the cryptographic approaches for confidential machine learning (CML), primarily on model training, while also covering other directions such as perturbation-based approaches and CML in the hardware-assisted computing environment. The discussion will take a holistic way to consider a rich context of the related threat models, security assumptions, design principles, and associated trade-offs amongst data utility, cost, and confidentiality. 
\end{abstract}



\end{abstractbox}
%
\end{fmbox}%

\end{frontmatter}

\newpage

\section{Introduction}\label{sec:intro}
Data-driven methods, e.g., machine learning and data mining, have become essential tools for numerous research and application domains. With abundant data, data owners can build complex analytic models for areas ranging from social networking, healthcare informatics, entertainment, and advanced science and technology. However, limited in-house resources, inadequate expertise, or collaborative/distributed processing needs force data owners (e.g., parties that collect and analyze user-generated data) to depend on somewhat untrusted platforms (e.g., cloud/edge service providers) for elastic storage and data processing. As a result, cloud services for data analytics, such as machine-learning-as-a-service (MLaaS), have been rapidly growing during the past few years. While untrusted platforms refer to all non-in-house resources not directly owned by the data owner, we will use cloud services to represent them here forth.

When outsourcing sensitive data (e.g., proprietary, human-related, or confidential data), data owners have raised concerns in privacy, confidentiality, and ownership \cite{sharma18ic,duncan12}. On the one hand, cloud users cannot verifiably prevent the cloud provider from accessing their data; i.e., in practice, using public clouds often means one must fully trust the cloud provider. On the other hand, public cloud providers are not immune to security attacks leading to sensitive data breaches. Recent security incidents, including insider attacks  \cite{chen10,duncan12} and external security breaches at the service providers \cite{mansfield15,unger15}, show the risks are aggravating by day. Researchers and practitioners have developed solutions to protect the confidentiality of cloud data at rest. For example, Google Cloud Platform has allowed users to include an external key manager to store encrypted data on the cloud with a third party (e.g., Fortanix) stores and manages keys off the cloud. However, it remains a critical challenge for data owners and cloud providers to protect confidentiality in computing, i.e., learning models on the cloud, while protecting the confidentiality of both the training data and the learned models. 

In the past few years, researchers have made some progress in developing novel confidential machine learning (CML) approaches for model training with encrypted data.   A successful CML approach is not straightforward. Unlike traditional machine learning approaches, a practical CML framework wrestles in balancing security (confidentiality) guarantees, costs, and model quality, while allocating appropriate workload distributions between cloud and client.  Direct application of cryptographic and privacy-protection methods such as fully homomorphic encryption (FHE) \cite{gentry09} and garbled circuits (GC) \cite{yao86} in a homogeneous fashion do not usually meet the criteria for practical CML approaches. Most efficient approaches have been using hybrid methods that combine multiple primitives instead of a homogeneous translation. Recent studies \cite{niko13,niko13sp,demmler15,mohassel17,sharma18tkde,sharma19} have followed this direction to effectively reduce performance bottlenecks and other practicality issues in developing CML solutions. However, the underlying techniques in these studies scatter among several papers making the basic principles are unclear. The purpose of this survey is to uncover these basic principles and accurately organize the existing techniques under a unified framework so that researchers and practitioners can quickly grasp the development and challenges in this new area of research.


\textbf{Contributions and Organization Overview.} 
Capturing a comprehensive view of a complex and new topic like confidential machine learning is challenging. We primarily focus on frameworks for \emph{model training} using cryptographic techniques that guarantee strong (semantic) security with practical cost overburden. A complete machine learning service usually includes a model application (or model inference) component that applies the learned model to generate a prediction for new input data, equivalent to secure function evaluation. The confidential model inference is much simpler and in a more mature state than confidential model training, therefore, not covered in this survey. Interested readers may refer to the related studies about confidential inference with pre-trained models, such as Gilad-Bachrach et al.  \cite{gilad16}, Bost et al. \cite{bost15},  Hesamifard et al. \cite{hesamifard17}, and Rouhani et al. \cite{rouhani18redcrypt}.

This survey paper presents a unified perspective on designing and implementing different CML model learning methods with state-of-the-art cryptographic approaches. Despite numerous machine learning methods \cite{hastie01}, the studies on CML methods have focused on only a few specific machine learning methods. On the other hand, researchers have applied several cryptographic methods to realize CML frameworks. We observe that many clever CML techniques apply to specific machine learning algorithms without clearly establishing the basic principles for extending these techniques to broader machine learning algorithms. To systematically understand the set of developed techniques in CML, we summarize them under a general framework, the decomposition-mapping-composition (DMC) procedure + design and selection of crypto-friendly algorithms. The DMC procedure involves: decomposing the target machine learning algorithm into several components, mapping these components to their cryptographic constructions, and finally composing the CML solution with the confidential component counterparts. Moreover, several CML approaches adopting the DMC process development have a unique additional feature: they use ``crypto-friendly'' alternative machine learning algorithms or components to achieve more efficient protocols. Keeping these observations in mind,  we develop a systemization framework to summarize the design principles, strategies, cryptographic techniques, and optimization measures, which have been applied to solve the challenging problems in confidentially learning models over encrypted data. 

We organize the survey based on underlying design principles of CML rather than any specific machine learning problems. As part of the survey, we summarize the experiences and learnings in each category of CML topics as \emph{insights} and \emph{gaps}. This work promotes practical aspects of applying cryptographic primitives in CML at their current level of maturity. Focuses will be on how different frameworks balance the associated trade-offs amongst cost, confidentiality, and data utility or model quality in different threat models and privacy settings. The survey, however, does not cover the orthogonal line of research that aims to optimize fully expressive primitives such as FHE and GC schemes. This survey will be a great resource for researchers to adopt and advance privacy-enhancing technologies in solving novel research questions and for practitioners to learn the best practices and avoid common pitfalls.

In the following sections, first, we will include the necessary background knowledge, notations, definitions, and the targeted threat model in Section \ref{sec:background}. Then, in Section \ref{sec:systemization}, we present the systematization framework along with the basic principles and methodologies in the CML development. After that, we briefly discuss the homogeneous approaches that aim to translate any machine learning algorithm into a confidential one with a single cryptographic primitive (Section \ref{sec:homo}). Next, we move to the main theme: the compositional hybrid approaches (Section \ref{sec:hybrid}), which have resulted in more efficient protocols for complex machine learning models. We will also cover several topics, such as security proofs and common evaluation methods for cryptographic protocols in Sections 7 and 8. Finally, we briefly review other non-cryptographic-protocol approaches, including the perturbation methods and hardware-assisted (e.g., SGX) methods. 

\section{Related Work} 
A few survey papers are related to the topic of this paper. Shan et al. \cite{shan18} focus on techniques for practical secure outsourced computation, using machine learning as a sample application. However, it does not comprehensively cover the major approaches as we do. 
Attacks on the integrity of machine learning models have also raised serious concerns due to the wide applications of machine learning in real-life scenarios such as self-driving cars \cite{sorin19}. Different from our survey focusing on the confidentiality of the model learning process, Papernot et al. \cite{papernot18} focus on the integrity of training data, learning process, models, and model application. 

There are also several survey papers on a specific category of cryptographic primitives. Since the first fully homomorphic encryption scheme was published in 2009 \cite{gentry09}, it has been an active research area during the past decade. Acar et al. \cite{acar18} have a comprehensive review about the current development of homomorphic encryption schemes. Secure multi-party computation methods, including the garbled circuits and secret sharing methods, have been actively developed for the past two decades. Readers may find more information from other sources \cite{lindell20,evans18}.  

Differentially private machine learning frameworks are somewhat related to CML but hold a distinct thread model that aims to share data and models. They assume that the data consumer (i.e., model developer or model users) is not trusted, who may try to reveal private information in the training data shared by data owners or data contributors. It does not protect the ownership of data and models as the purpose is to share them without breaching individuals' privacy in the training data. Along with recent developments on differentially private deep learning such as Abadi et al. ~\cite{abadi16} and Shokri et al. \cite{shokri15},  Ji et al. \cite{ji14} and Sarwate and Chaudhari \cite{sarwate13} also provide excellent surveys on this topic.  Other studies in privacy-preserving data mining (PPDM)  \cite{charu10book,matwin13,aldeen15,sachan13}  also aim to share the data (and the models) while preserving individual's privacy, thus excluded from our survey. 

\section{Preliminaries of CML Approaches} \label{sec:background}
In this section, we review the terms and concepts used in the literature. First, we look at the representative system architectures considered in the published confidential machine learning (CML) approaches based on cryptographic protocols. Then, we examine how different threat models, associated confidential assets, and considered attacks affect CML designs. Finally, we briefly describe prevailing cryptographic and privacy primitives that serve as the skeleton of most CML approaches. 

\subsection{System Architectures} \label{sec:system_arch}
The CML research is motivated by the cloud computing paradigm and then extended to more scenarios, such as edge computing and services computing. Thus, we use ``Cloud'' as the representative of untrusted platforms in CML system architectures henceforth. Such a system may involve cloud providers, optional cryptographic service providers, data owners or application service providers, and data and model consumers. 
Figure \ref{fig:hbc_CP} shows an architecture with a data owner outsourcing its data and computation to a single cloud provider. The data owner must ensure the cloud provider does not compromise any proprietary and privacy-sensitive data. A few homomorphic-encryption-based frameworks, e.g., Graepel et al. \cite{graepel12} and Lu et al. \cite{lu17}, present protocols for training machine learning models over encrypted data outsourced to a cloud provider without almost any engagement of the data owner. However, the associated cost makes these protocols unrealistic in real-life scenarios. An alternate strategy would involve the data owner in minimal tasks intermediately to simplify the single cloud architecture framework \cite{sharma18tkde}. As long as the cloud takes the majority of the workload and the client's cost is practical, e.g., linear or sublinear to the number of records, more efficient protocols can be possible. 

As some protocols become too expensive for the data owner to assist cloud-centric learning, the architecture was evolved to a multi-server(cloud) setting. A data owner may choose to rely on two or more cloud providers to reduce the overall expense of learning.  The second party may be as equally capable as the first party \cite{mohassel17}, or in the case of a cryptographic service provider (CSP), which manages keys and assists the cloud with intermediate decryption operations and light-weight computations \cite{niko13,niko13sp,sharma19}. The two non-client parties in such an architecture carry out secure multi-party computations without any of the parties learning the training data and the trained model. This setting also assumes that the two parties do not collude with each other, thus slightly more vulnerable than the client-cloud two-party setting. Figure \ref{fig:parties} shows such a framework that uses a garbled circuit.

\subsection{Threat Models}\label{sec:threat_models} 
In this section, we examine the widely accepted threat models in the context of CML. We focus on the following aspects: the assumptions on the adversaries and the related confidential assets in CML.

\textbf{Assumptions on Adversaries.}
Most CML approaches \cite{sharma19,mohassel17,niko13,niko13sp,graepel12} adopt the honest-but-curious (or semi-honest) adversary model to describe the untrusted cloud provider. Honest-but-curious parties, by definition, perform their share of tasks obediently, i.e., guarantee data and model integrity and follow the pre-defined protocols exactly. However, they might clandestinely snoop the storage, interactions, and computations to learn private information. Data owners and data contributors' concerns about data and model leakages, even when the infrastructure platforms are reputed, are alleviated by preserving the confidentiality of data and models. Many CML approaches also use an honest-but-curious cryptographic service provider to design more efficient protocols.   

Some CML approaches additionally address an adversary which actively seeks to compromise data and model confidentiality by performing additional probing tasks, e.g., by inserting crafted records or secretively running the algorithms on a selected record set offline. Sharma and Chen \cite{sharma19} address the possibility of an adversary who may actively track identifiable training records to the datasets and follow the computations to infer the information about other training records. Nikolaenko et al. \cite{niko13sp} consider an adversary that selectively runs the machine learning protocol over an individual's data to draw personal inferences from the learned models. 

Nevertheless, with either passive or active adversaries, CML approaches assume that the data and model integrity not be compromised at the end of the training. This assumption distinguishes CML from other studies such as attacks on machine learning by polluting training data or modifying learned models \cite{liu_adv18}. 



Moreover, the CML approaches often assume non-collusion between the involved parties, for example, between the cloud provider and the CSP \cite{niko13,niko13sp,sharma19} or the two cloud providers \cite{mohassel17} in the two-server architecture. Collusion between the two parties in these frameworks directly compromises the privacy of the training data and learned models. 

Most CML approaches assume that data and model consumers are trusted, which is orthogonal to the applications of differential privacy \cite{shokri15,abadi16} that specifically targets untrusted data and model consumers. Furthermore, CML approaches assume properly secured infrastructures and communication channels to exclude external attacks and focus on the CML-specific challenges. 

\textbf{Confidential Assets at Risk.}
An adversarial party may be interested in the confidentiality of \textit{sensitive data} and the \textit{generated models}. All CML methods protect the training data feature vectors. Some methods designed for supervised learning \cite{graepel12,niko13sp} expose the training data labels to simplify their secure modeling algorithms with the assumption that knowing the labels will not bring significantly more information to adversaries, which might be false for some applications. Some CML studies also expose unprotected models \cite{graepel12,niko13sp,lu17}. However, recent studies  \cite{fredrikson14,fredrikson15,hitaj17,shokri16,song20} have shown that an adversary may use crafted data to infer sensitive training data or use the advanced features in deep learning models to breach data privacy. Furthermore, the intermediate results of outsourcing computations in the setting of federated learning, for example, the intermediate representation in a convolutional neural network learning, may reveal information about the private training data \cite{shokri15}. Thus, CML must protect both data and model confidentiality.

\subsection{Cryptographic Primitives }\label{sec:primitives}
The cryptographic primitives are the fundamental building blocks for CML approaches. Some of these primitives are more expressive -- meaning they can implement more types of functions or higher-level functions. On the other hand, some primitives are more cost-efficient than others. To make this survey self-contained, in this section, we briefly cover the most frequently-used primitives in existing CML approaches.

\textbf{Additive Homomorphic Encryption (AHE).} AHE schemes  (e.g., Paillier encryption \cite{paillier99}) allow the additive operation over encrypted messages without decryption. For any two integers $\alpha$ and $\beta$, an AHE scheme allows the additive homomorphic operation: $E(\alpha+\beta) = f(E(\alpha), E(\beta))$ where the function $f$ works on encrypted values. Conceptually, with one of the operands unencrypted, a ``pseudo-homomorphic'' multiplication between two messages can be expressed as a series of additions\footnote{Some methods like Paillier encryption \cite{paillier99} allow more efficient pseudo-homomorphic multiplication.}, i.e.,$E(\alpha\beta) = E(\sum_{i=1}^{\beta}\alpha)$. With homomorphic addition and pseudo-homomorphic multiplication, one can derive pseudo-homomorphic dot-product of vectors, matrix-vector multiplication, and matrix-matrix multiplication. However, the unencrypted operands in these operations either need to be non-sensitive information or protected with some masking and de-masking mechanism \cite{sharma18tkde,sharma19}. ElGamal, Goldwasser-Micali, Benaloh, and Okamoto-Uchiyama cryptosystems are some additional examples of AHE schemes \cite{acar18}.
 
\textbf{Somewhat Homomorphic Encryption (SHE).}  There are many encryption schemes in this category (e.g., BV, BGV, NTRU, GSW, BFV, and BGN \cite{acar18} and their variations such as TFHE \cite{chillotti20} and CKK \cite{hee17}). SHE schemes allow both homomorphic additions and multiplications over encrypted messages, while the number of consecutive multiplications is limited to a few. A popular SHE scheme used in CML is the ring learning-with-error (RLWE) scheme that relies on the intractability of the learning-with-errors (LWE) problem on polynomial rings \cite{BGV12}. Theoretically, RLWE supports arbitrary levels of multiplications. Therefore, it is considered to be fully homomorphic. However, due to the associated high cost for deeper levels of multiplications,  RLWE is more suitable as a SHE scheme only (i.e., 1-3 levels of multiplications).  A ciphertext in RLWE is represented as a two-tuple $(c_0, c_1)$, where $c_0$ and $c_1$ are polynomials. Let $C_i=(c_{0,i}, c_{1,i})$ and $C_j$ be the ciphertext of any two values. The encrypted addition of the two values is simply $(c_{0,i}+c_{0,j}, c_{1,i}+c_{1,j})$. The encrypted multiplication is translated to a series of polynomial operations on the ciphertext elements. RLWE allows multiple levels of multiplication at a certain cost. For details, please refer to the paper \cite{BGV12}. \emph{Message packing} \cite{BGV12} enables packing multiple ciphertexts into one polynomial, which considerably reduces RLWE's ciphertext size and optimizes linear algebra operations \cite{halevi14algo}. HELib library \cite{halevi14algo}  is a popular implementation of the RLWE scheme.

\textbf{Garbled Circuits (GC).} Garbled Circuits (GC) \cite{yao86} allow two parties, each holding an input to a function, to securely evaluate a function without revealing any information about the respective inputs. GC can express arbitrary functions using several basic gates such as AND and XOR gates in a secure two-party computation setting (usually with a Cryptographic Service Provider (CSP)). One party constructs the circuit, whereas the other evaluates it. Despite several GC cost optimization techniques, such as Free XOR gates \cite{kolesnikov08}, Half AND gates \cite{zahur15}, and OTExtensions \cite{asharov13}, GC still incurs high communication costs. Therefore, one must carefully examine its use in composing CML frameworks. FastGC \cite{huang11} and ObliVM \cite{liu15} are two popular GC libraries. 

\textbf{Randomized Secret Sharing (SecSh).} The randomized secret sharing method \cite{demmler15} protects data by splitting it into two (or multiple) random additive shares outsourced to two (or more) non-colluding untrusted parties. The two parties compute on the respective shares and return the results also as random shares. Addition is straightforward as $\alpha + \beta = (\alpha_0+\beta_0) + (\alpha_1 + \beta_1)$ with $\alpha$ and $\beta$ distributed between two parties 0 and 1. Multiplication, however, is expensive as it depends on the beaver triplet generation method \cite{demmler15,mohassel17}, which further depends on expensive AHE or Oblivious Transfer (OT) schemes to exchange the intermediate results securely. 

\textbf{Random Additive Masking.} 
A data owner may generate a random mask to hide the sensitive data, which will be stripped off at a certain step in the CML protocol to recover the desired result. Due to its low cost, it frequently serves as an auxiliary tool for a complex protocol, for instance, in CML for spectral clustering \cite{sharma18tkde}, boosting \cite{sharma19}, and matrix factorization \cite{niko13}.

\section{Systematization Framework} \label{sec:systemization}
It is challenging to have a clear understanding of the whole body of CML model training methods due to the following reasons. First, the number of machine learning models is huge \cite{hastie01} and even the most used ones are around tens \cite{wu07}. They are so different that no unified framework can be used to describe them. Second, security researchers are often more interested in a specific utility-preserving cryptographic primitive method and pick the machine learning algorithms they are most familiar with. As a result, the results are scattered with focuses on either a specific machine learning model or the application of a novel cryptographic primitive. There is no thorough understanding of which primitive method (or framework) is best for a specific machine learning method or whether a CML method can be extended to other machine learning models. The fundamental principles are missing for solving all (or most) CML model training methods. 

\textbf{Categories of CML approaches.} We believe this survey is the first effort to systematically organize and analyze the whole body of most representative CML approaches. We focus on the major category of methods: the pure software-based \emph{cryptographic protocols}, while also briefly reviewing the \emph{perturbation-based} approaches and the \emph{hardware-assisted} approaches. Figure \ref{fig:system} shows the systematization framework. The fundamental features of the three categories are as follows.  
\begin{itemize}

\item The cryptographic protocols are the focus of this survey, which can be further divided into two categories: those using one cryptographic primitive homogeneously and those employing novel hybrid compositions of multiple primitives. The homogeneous approaches take one of the homomorphic encryption (HE) schemes or garbled circuits to develop the solution. The hybrid approaches involve multiple primitives and often a clever composition strategy to achieve lower overall costs. We will analyze them in more detail.

\item The perturbation-based CML approaches depend on novel data transformations to preserve a certain type of data utility, e.g., Euclidean distance, that is critical to one or multiple machine learning methods. Their security mainly depends on secret transformation parameters and random noise addition, holding a different and somewhat weaker security notion compared to cryptographic protocols. However, they are often much more efficient and thus appealing for many applications that seek better protection than plaintext-based approaches while not taking significantly more overhead.  

\item The third category depends on \emph{trusted execution environment}, such as Intel SGX \cite{sgx-explained}, which demands hardware-level supports and are thus distinct from the former two categories of pure software approaches. The hardware-level features enforce \emph{secure enclaves}, in which the adversaries cannot observe the running programs and data. 
\end{itemize}

\textbf{Common CML Development Strategies.} We look into a unified framework to analyze both the homogeneous and hybrid approaches. Fundamentally, most approaches aim to design an efficient and secure transformation of the specific (or a class of) machine learning algorithms for the setting of two or three distributed parties (see Section \ref{sec:system_arch}).  To make the transformation easier, researchers often implicitly use the Decomposition-Mapping-Composition (DMC) procedure: decomposing the target algorithm into different subcomponents, mapping the sub-components to crypto-primitives, and composing the CML framework with the confidential sub-components.  Many approaches skip the description of this whole procedure and only present the final composition, which creates difficulties for newcomers to fully appreciate the fundamental ideas scattered in several approaches. 

Beyond the straightforward DMC procedure, we have also noticed a unique feature \cite{sharma19} specific to the CML development: finding ``crypto-friendly'' alternative machine learning algorithms or components. This feature is unique to machine learning algorithms because all machine learning algorithms essentially try to find an approximate model fitting the training data, and there is no unique model for a specific problem, only better or worse ones.  In general, machine learning methods can be roughly categorized into two types: supervised learning that depends on labeled datasets and unsupervised learning \cite{hastie01}. For each type, there are numerous algorithms working under the same setting but performing differently for specific applications or datasets. Even for the same algorithm, there are many variants. For example, different base classifiers can be used to make ensemble classifiers \cite{schapire99}, and different activation functions can be used for neural networks \cite{lecun15}. Among so many machine learning algorithms, some are more crypto-friendly, i.e., they can be converted to more efficient CML solutions.

With all these features in mind, we reassemble the common development framework behind most CML approaches (Procedure \ref{pro:cml-procedure}). 
\floatname{algorithm}{Procedure}
\begin{algorithm}[h!]
\caption{A common procedure for developing CML methods}\label{pro:cml-procedure}
\begin{algorithmic}[1]
\Procedure{Generalized procedure for CML development}{$A$}     
    \State $A$: the target algorithm
    \State Identify the desired architecture and involved parties.
    \State Identify a list of alternative algorithms of $A$
    \For{Each candidate algorithm}  
        \State decompose the algorithm to basic components
        \For{Each component}
        	\State identify possible approximate/equivalent solutions
        	\For{Each solution}
        	\State identify candidate crypto-primitive mappings
        	\EndFor
        	\State select the best solution and mapping. 
        \EndFor
        \State find the best composition method.
    \EndFor 
    \State evaluate the candidate alternative algorithms and identify the best one.
\EndProcedure
\end{algorithmic}
\end{algorithm}

Note that most of the steps in this procedure cannot be automated, and thus each specific approach represents a result of enormous efforts behind the scene. Next, we analyze the homogeneous and hybrid approaches under this unified procedure. 

\section{Homogeneous Cryptographic Approaches} \label{sec:homo}
Homogeneous approaches rely on a single primitive to construct the framework protocols. The primitives used in the homogeneous composition of CML are broadly in two categories: (1) Fully Homomorphic Encryption (FHE)  and Garbled Circuits (GC) and (2) Additively Homomorphic Encryption (AHE) and Somewhat Homomorphic Encryption (SHE). Since FHE implements arbitrary levels of homomorphic addition and multiplication and GC implements the boolean gates, in theory, they can individually construct all CML algorithms. FHE and GC are, therefore, the most expressive privacy primitives. However, both FHE and GC are too expensive to be practical when mapped to for training complex CML models. Oppositely, AHE and SHE schemes provide limited support for encrypted operations, therefore, less expressive and can only enable relatively simple algorithms. Most approaches we discuss next are relatively simple, and thus AHE or SHE scheme is sufficient. The decomposition and mapping steps of the DMC procedure described in the last section are still at play in the homogeneous approaches, but the composition step is trivial. 

AHE and SHE are widely used to construct homogeneous solutions for applications involving only one or a few multiplications, including the elementary statistical aggregation functions, such as average, sum, and variance. Graepel et al. \cite{graepel12} present a SHE-based framework for learning Fisher's linear discriminant analysis and Linear Means Classifier models on encrypted data. However, the implemented models are limited to linearly separable datasets. Lu et al. \cite{lu17} apply SHE for more sophisticated principal component analysis, and linear regression training \cite{hastie01}. However, due to the limited message space of the selected SHE implementation (60-bits in HELib) and the limited number of possible multiplications, only low data dimensionality (about 20) and a few training iterations were used in their evaluation. Such restrictions, however, resulted in only sub-optimal models. 

More sophisticated machine learning algorithms often result in expensive homogeneous solutions. Phong et al. \cite{phong18} employ LWE and Paillier encryption in encrypting the gradients in their privacy-preserving deep learning framework. The framework, however, takes over 2.5 hours to complete one iteration of a simple neural network training for 20,000 MNIST images. Researchers also aim to provide libraries for homogenous learning based on Garbled Circuits (GC). However, their uses are limited in practicality due to huge costs \cite{liu15}. Liu et al.  \cite{liu15} present a GC-based KMeans learning framework that involves two untrusted servers. The associated cost overburden, however, is far from efficient in real-world settings. For example, the KMeans implementation required over 2,000 million AND gates and more than 200 GB communication for clustering just 6,000 data points. Rouhani et al. \cite{rouhani17} propose a deep learning model inference frameworks using garbled circuits to protect both the model's parameters and test data samples. Similarly, the costs are staggeringly high.

\emph{\textbf{Insight.} Homogeneous solutions are often limited to simple functions involving only additions (for AHE), a few multiplications (SHE), or a few comparisons (GC). Individually, these crypto primitives are not practical to construct complex CML algorithms. However, they can be valuable components for hybrid solutions, as we will see later.} 

\section{Hybrid Composition}\label{sec:hybrid}
As discussed above, depending on a single cryptographic primitive to compose a sophisticated CML algorithm is impractical. However, each primitive has its unique strengths and shortcomings (e.g., performance, storage, bandwidth advantage) in attaining certain operations. This realization leads to an interesting strategy: can we combine different primitives in such a manner to compose secure yet more optimized protocols? The idea of hybrid composition is thus, mixing and switching amongst several privacy primitives to avoid the associated cost bottlenecks and restrictions of any individual primitive.

This section will look into the details of specific steps of the DMC procedure. First, we dissect the common sub-components and underlying operations in machine learning algorithms. We examine the various ways to implement these sub-components and operations confidentially.  Then, we explore the different switching and mixing strategies, including some recent automated ones, essential to hybrid CML frameworks in practice. Finally, we discuss the unique feature or desired requirement of CML development: designing crypto-friendly machine learning algorithms or sub-components for cost-efficient and practical CML solutions. 

\subsection{Basic Operations}\label{subsec:components}
We devote this subsection to inspecting the mapping of the foundational sub-components of the target machine learning algorithms to their confidential versions. We observe that some of these mappings are practical or crypto-friendly, whereas others may face cost bottlenecks and limitations. The understanding of the different implementations of basic operations will affect the composition strategies. 

\textbf{Simple Arithmetic Operations}\label{subsec:arithmetic}
With AHE or a SHE encryption scheme, one can conveniently add two encrypted integers. Adding two $b$ bit integers with the Paillier cryptosystem involves modular multiplication with O($b^2$) complexity. Additions with an RLWE-like scheme involve polynomial additions linear to the number of bits for the given polynomial degree \cite{chakarov19}. With a specific integer encoding, subtraction becomes trivial expressed as encrypted additions. SHE schemes allow homomorphic multiplications over encrypted integers. RLWE-like crypto-systems allow several rounds of multiplications and additions.  However, with each additional multiplications, the ciphertext noise, cipher size, and cost increase. Generally, multiplying two $b$ bit integers with RLWE-like crypto-systems involves homomorphically computing O($b^2$) AND circuits \cite{chakarov19}. On the other hand, the AHE scheme requires one of the operands to be unencrypted to realize multiplication expressed as summations. With Paillier encryption, multiplication is modular exponentiation of encrypted $b$-bit message by the unencrypted $b$-bit operand with a cost complexity of O($b^3$). The only caveat of using AHE-multiplication is that if the unencrypted operand is privacy-sensitive, a mechanism to mask it needs to be augmented, the masking recoverable after the multiplication is complete \cite{sharma18tkde, niko13}. 

Additions and subtractions are trivial with randomized secret sharing in the multi-party setting with constant time complexity. Each party performs additions and subtractions on respective shares of data and shares the results for recovery. A GC protocol for addition requires two parties to construct $O(b)$ many AND gates and carry out  $O(b)$ communication, encryptions, and decryptions along with $O(b)$ oblivious transfers when adding two $b$ bit integers. Multiplication with randomized secret sharing involves a costly multiplicative triplet generation scheme that relies on Oblivious transfer or AHE \cite{demmler15, mohassel17}. For example, the AHE-based scheme incurs transmission of two encrypted integers between the parties and performing two homomorphic encryptions, multiplications, additions, and decryptions by each party. Multiplying two integers of $b$ bits with GC, on the other hand, requires construction and evaluation of  $O(b^2)$ AND gates involving $O(b^2)$ communication, encryption, and decryption.

\textbf{Comparison.}
Comparison is essential in many operations, such as sorting vectors and applying activation functions in training neural networks. Unfortunately, comparing two encrypted or protected integers is not trivial. Graepel et al. \cite{graepel12} pose the complexity of comparison as the reason to avoid algorithms like perceptrons and logistic regression in their SHE-based confidential ML framework. Veugen \cite{veugen} presents a client-server interactive comparison protocol for two encrypted integers based on the AHE scheme, which involves computation and transfer of $b$ many AHE encrypted bits. Each comparison incurs $O(b)$ homomorphic multiplications for both client and server. Lu et al. \cite{lu17} use the technique of  ``greater than'' protocol \cite{golle06} optimized with the message packing of the RLWE scheme for comparing two encrypted messages in a two-party setting. However, the associated complexity is an astonishing $O(2^b/h)$ of homomorphic additions when comparing two $b$-bit integers while packing $h$ messages in a ciphertext. With GC, a comparison between two $b$-bit integers is possible with $O(b)$ AND gates and $O(b)$ communication, encryption, and decryption by two parties. Since GC-based comparison for full integers is expensive, one may use an efficient one-bit sign checking protocol \cite{mohassel17,sharma19} by encoding negative integers as two's complement, making the comparison cost is constant to the number of bits. Note that the GMW protocol of Goldreich, Micali, and Wigderson \cite{goldreich87} can perform comparisons just as garbled circuits but with $O(b)$ rounds. A similar sign-checking protocol is possible with GMW. However, the GC-based comparison seems the popular choice in current solutions.

\textbf{Division.}
Division can be essential to many analytics algorithms, e.g., from the computation of mean to the implementation of complex algorithms such as K-means \cite{bunn07} and Levenshtein distance \cite{rane10}. Despite its prevalence and importance, translating division to its confidential version is expensive and often results in a performance bottleneck \cite{lazzeretti11}. Veugen \cite{veugen} presents a protocol for exact division in a client-server scenario, using the AHE  scheme and additive noise masking. However, the protocol requires the divisor to be public knowledge. On top of that, the protocol requires $O(b)$ homomorphic comparisons and $O(b)$ encrypted communication for division between two $b$-bit integers. Dahl, Chao, and Tomas \cite{dahl12} present two AHE-based division schemes that rely on Taylor approximation in a secure multi-party setting. The schemes brought expensive $O(b)$ encrypted communication. It is possible to perform integer divisions with GC when the two parties hold the numerator and denominator respectively in a 2-party setting \cite{lazzeretti11,niko13sp}. However, even with several optimizations, a division between two $b$-bit integers involves the construction and evaluation of a circuit with $O(b)$ non-XOR gates \cite{lazzeretti11}. A more practical solution would be to decrypt the operands at a crypto-service provider and conduct division on plaintext before finally encrypting the result.

\textbf{Linear Algebra Operations.}
Linear algebra operations, such as vector dot products, matrix-vector multiplication, and matrix-matrix multiplications, are the core operations for many machine learning algorithms. They are commonly implemented with the cryptographic versions of additions and multiplications with some tricks in  RLWE-based SHE for improved efficiency. Among all available methods, the AHE and SHE-based implementations are the most efficient ones. 

A dot product $x_{k}^Ty_{k}$ involves $O(k)$ element-wise homomorphic multiplications and additions. Similarly, a matrix-vector multiplication $A_{n \times k}x_{k}$ involves $O(nk)$ homomorphic multiplications and additions, and a matrix-matrix multiplication $A_{n\times k}B_{k\times m}$ involves $O(nkm)$ multiplications and additions. With the AHE scheme, one of the operands must remain unencrypted for these multiplicative operations. Therefore, the unencrypted operand needs some level of protection, e.g., novel randomized masking \cite{sharma18tkde} with a minimized cost. With the message packing feature for the RLWE-like SHE scheme, one can easily vectorize the vector and matrix operations with message packing to gain more efficiency \cite{halevi14algo}. With such facilities, Jiang et al. \cite{jiang18} can optimize matrix-matrix multiplication with only $O(k)$ complexity for symmetric matrices of $k$ dimensions. 

Randomized secret sharing enables linear algebraic operations with the multiplicative triplet generation approach in a multi-party setting. However, this involves the expensive AHE or OT-based multiplicative triplet generation schemes as used in \cite{mohassel17,demmler15}. In computing a matrix-vector multiplication $Ab$, each party is responsible for $O(n+k)$ encryptions and upload, $O(nk)$ homomorphic multiplications, $O(nk+n)$ homomorphic additions, and $O(n)$ decryptions. 

One can easily map linear algebra operations to garbled circuits.  GC-based vector and matrix addition/subtraction require $O(kb)$ and $O(nkb)$ AND gates where $b$ is the number of bits in the vector and matrix elements. They also result in $O(kb)$ and $O(nkb)$ communication, encryption, and decryption operations, respectively. GC-based dot product for two $b$ bit vectors with $k$ dimensions is a collection of sub-circuits for multiplication and additions, which consist of $O(kb^2)$ AND gates. The cost also involves $O(b^2)$ encryption and decryption, and $O(b^2)$ encrypted communication. The GC-based dot product can easily extend to matrix-vector and matrix-matrix multiplication. However, GC-based linear algebra solutions are more expensive than HE-based ones.

\subsubsection{Empirical Cost Comparison} 
We have formally analyzed different crypto implementations for each of the major operations. However, some of them look close in terms of bigO complexity levels. To have a better idea how the cost differences look like for the different implementations of the same operator, we also prepare Table \ref{tab:arithmetic_cmp}. Since this comparison rests on a specific hardware configuration and software implementation, readers should only focus on the relative differences rather than the actual numbers. After a careful study of available AHE and SHE implementations, we choose the most efficient one for each category: we use the HELib library \cite{halevi13} for the RLWE encryption scheme and implement the Paillier cryptosystem \cite{paillier99} for the AHE encryption scheme. We adopt the ObliVM (oblivm.com) library for the garbled circuits. We also take the AHE scheme for the multiplicative triplet generation when using the randomized secret sharing (SecSh) method. We pick cryptographic parameters\footnote{The Paillier cryptosystem uses a 2048-bit key size. We set the degree of the corresponding cyclotomic polynomial in the RLWE scheme to $\phi(m)$ = 12, 000 and c = 7 modulus switching matrices, which gives us h = 600 slots for message packing.}corresponding to $112$-bit security. All schemes allow at least 32-bit messages-space overall. The RLWE parameters allow one full vector replication and at least two levels of multiplication. Note that the GC and SecSh costs are for the two-party setting, which has to involve communication costs between the two parties. Thus, we also include the bytes of exchanged messages for these methods. We run the experiments on an Intel i7-4790K CPU running at 4.0 GHz using 32 GB RAM with Ubuntu 18.04. 

Table ~\ref{tab:arithmetic_cmp} compares the related costs of arithmetic operations over integers. We have observed that the AHE scheme has the most efficient arithmetic additions and multiplications. However, for comparison and division, the 2-party garbled circuits are the only viable option. The table also shows the costs for the linear algebraic operations. The observation is consistent with the simpler arithmetic operation of additions and multiplications. As we can fit multiple messages in a ciphertext when using the RLWE scheme, the vectorized additions and multiplications are much more efficient than the non-vectorized additions and multiplications. The RLWE with message packing realizes homomorphic additions more efficiently when compared to the Paillier scheme. The RLWE costs for dot product and matrix-vector multiplication involve the ciphertext replication costs. Although better than without message packing, the RLWE scheme with the vectorized linear algebraic operation is still slower than the Paillier solutions.  Randomized secret sharing is almost free for vector addition but involves higher computation and communication costs for the dot product and matrix-vector multiplication. Garbled circuits appear to be the worst solution for the confidential versions of the linear algebraic operation with higher computation and communication costs between the two parties. Although the Paillier implementation shows performance advantages over RLWE on arithmetic operations,  it requires one operand to be plaintext.  Paillier's encryption and decryption costs, however, are higher than that of RLWE \cite{sharma18tkde}. When CSP is involved in a solution, encryption and decryption costs will become a critical performance factor. These cost comparisons on the basic operations will be useful for readers to analyze and compare a pair of CML protocols, especially when not all CML methods are open-source. 

We do not experimentally compare complete CML approaches because 1) different approaches often solve different ML problems, which makes the comparison difficult, and 2) not all approaches have open-sourced their implementation or shared executable binaries. However, we hope the empirical comparison between different implementations for basic operators gives an intuitive understanding of the rationales behind different CML design strategies and optimization methods. We refer readers to the papers describing CML approaches that often contain detailed performance comparisons between selected CML approaches.

\emph{\textbf{Insight.}
Based on most studies, the most efficient constructions for confidential comparison are GC-based, while SHE and AHE are better candidates for linear algebra operations. Since most division schemes are too expensive, one should consider transforming the functions/algorithms with divisions to the equivalent (often approximately) ones that involve no division. 
}

\subsection{Switching and Composing Strategies} \label{sec:composition} 
When composing the confidential versions of operations implemented with different primitives, there is an important step: switching computation flows between the primitives. This switching often requires a second party in the CML frameworks, i.e., either the data owner, the second non-colluding cloud, or a CSP to achieve better performance. 

\textbf{HE to/from GC.} Switching from a HE component to a GC component involves a second server (e.g., a CSP) in the framework.  A straightforward approach would be including a data decryption circuit inside a garbled circuit to be evaluated by the two parties. However, such an approach is super-expensive \cite{niko13}. A more practical strategy \cite{niko13,niko13sp,sharma19} is to have the party holding the encrypted data, denoted $P_A$, mask it homomorphically before sending it to the second party, $P_B$ for decryption. The second party constructs the desired garbled circuit, where the first step of the garbled circuit is de-masking the data with inputs: the decrypted masked data from $P_B$ and the mask from $P_A$. 

\textbf{SecSh to/from GC.} Switching from a SecSh component to a GC component is straightforward in a two-party architecture. The two random shares in possession of the two parties can be their respective private inputs to the desired garbled circuits \cite{mohassel17,sharma19, riazi18}. Similarly, switching from GC to SecSh involves evaluating the GC and randomly distributing the output to two parties \cite{riazi18}.

\textbf{SecSh to/from HE.} A switch from randomized secret sharing to a HE component needs two involved parties to encrypt their respective shares.  Then, one of the parties homomorphically reconstructs the protected value from the shares. Similarly, a switch from a HE component to a randomized secret sharing protocol includes a masking mechanism (homomorphic noise addition) similar to the HE-to-GC switch discussed above. These two switches are relevant in the AHE-based multiplicative triplet generation protocol for randomized secret sharing \cite{mohassel17,demmler15}. 

Table \ref{tab:switch_cmp} provides some examples of switching between cryptographic primitives in well-known CML approaches. These switchings lead to simplification of the CML framework and cost optimizations, as explained in the ``Justification'' column of the table.
The ABY framework \cite{demmler15} covers different adapter-like switching protocols for the multi-party computation settings, where two servers hold the training data as arithmetic, boolean, or Yao's garbled shares. The ABY3 \cite{mohassel18_aby3} and BLAZE\cite{patra20} framework extend the switches to 3-party scenarios. These works, however, do not cover the switching from and to the homomorphic encryption schemes.

\textbf{Manual vs. Automated Composition.} 
Most existing CML approaches using the hybrid composition strategy \cite{mohassel17,sharma19,sharma18tkde,niko13sp} are manually composed as there are myriads of problem-specific details to address. A line of research explores the possibility of automatically composing the CML frameworks \cite{dreier11,henecka10}. Although promising, the automatic composition strategy of Dreier and Kerschbaum \cite{dreier11} depends on the availability of an extensive performance matrix for the different confidential versions of the target algorithms' components. Henecka et al. \cite{henecka10} propose the TASTY compiler that automatically compiles a given machine learning problem as a mixture of garbled circuits and homomorphic encryption in a secure two-party computation framework. However, the process is still not fully automated - it requires a privacy expert to design and specify the components as well as the recommended mappings. 

\emph{\textbf{Gap.} Due to the high complexity of formulating the component-wise costs and profiling the switching costs, the automated composition approaches are not yet fully mature. More importantly, as we will see in the next section, the construction of a practical CML solution involves one more crucial step that automated composition methods cannot help much. One must establish an in-depth understanding and analysis of the target ML algorithm to redesign a ``crypto-friendly'' algorithm. 
}

\subsection{Crypto-friendly ML Algorithms}\label{sec:crypto_friendly} 
So far, the DMC framework seems straightforward:  one decomposes the target machine-learning algorithm to its sub-components and maps them to cryptographic constructions, and the final composition becomes almost trivial except that the primitive switching requires some clever steps. With enough experimentation, one can find an optimal set of confidential components for the target ML algorithm. However, this straightforward strategy may only work for some problems.  Despite the best optimization of mapping and composition, one may still end up with an impractical protocol, although better than the homogeneous or other suboptimal compositions. The fundamental reason is that the original machine learning algorithms do not account for confidential computation. They are optimized to achieve the best model prediction power rather than to be crypto-friendly.  On the other hand, a less-known slightly-under-performing ML algorithm that attains the same learning goal might be more cost-effective to translate to its confidential version. Thus, an advanced design step critical to the DMC procedure is replacing or redesigning some of the underlying ML components or even the entire ML algorithm to find the most efficient CML protocols. 
Table \ref{tab:simplifying_alogo} summarizes some example CML frameworks that incorporate strategies to make their protocols crypto-friendly and hence more cost-effective.  Mohassel et al. \cite{mohassel17}, in their SecureML work, substitute the expensive softmax operation involving inverses with a ReLU-based function involving only one division. This way, the framework significantly reduces the cost bottlenecks in their protocol. Graepel et al. \cite{graepel12} cleverly avoid division of encrypted data in the framework for confidential linear means classifier and Fisher's linear discriminant analysis by replacing divisions with a multiplicative factor. Nikolaenko et al. \cite{niko13sp} use the more efficient Cholesky's decomposition instead of the expensive LU decomposition in solving a system of linear equations in their linear regression framework. Similarly, Nikolaenko et al. \cite{niko13} adopt the sorting-based matrix factorization solution to reduce the overall complexity of computing gradient descent with Cholesky's decomposition-based matrix factorization. Sharma and Chen \cite{sharma19} propose to train a boosting classifier over encrypted data with an ensemble of random linear classifiers (RLC) instead of decision stumps.  An RLC  takes mere $N$ encrypted comparisons, whereas a decision stump takes far too many comparisons. Naehrig et al. \cite{naehrig11} replace the exponential function (the sigmoid) in their logistic regression protocol with the Taylor approximation of exponentiation. Computing the exact exponential function would have led to the computation of many levels of multiplications over the encrypted message -- which would have been intolerably expensive with SHE schemes. Similarly, Sharma et al. \cite{sharma18tkde} replace the inherently expensive eigendecomposition $O(N^3)$ with cheaper $O(N^2)$ approximation algorithms of Lancozs and  Nystrom in their spectral clustering framework. 

Data reduction techniques such as subsampling and preserving the sparsity of matrix are also critical to performance. Nikolaenko et al. \cite{niko13}, in their matrix factorization framework, use a sorting network that optimizes the garbled circuit-based gradient descent algorithm by only updating it for the user ratings that are present in the training dataset.  Similarly, Sharma et al. \cite{sharma18tkde} propose a differential privacy-based graph submission mechanism that reduced total storage by over 15 times and costs involving encryptions and the associated homomorphic operations by over 20 times on the graph drastically when running the secure Nystrom method for spectral clustering. To sum up, although the approximate algorithms introduce some degradation to the learned models, they deliver desired cost practicality justifying the tolerable quality sacrifice. 


\emph{\textbf{Insight.}
For the same learning problem, there are numerous algorithms.  Even for the same learning algorithm, there are many variants \cite{hastie01}. The search space for optimal composition can be quite large. More difficultly, most well-known ML algorithms are best known for model quality or learning efficiency and none specifically designed with optimal CML in mind. Even worse, some crypto-friendly alternatives might have been forgotten or become obsolete due to their suboptimal quality or efficiency. The design of a good CML solution heavily depends on the designer's deep understanding of the ML algorithms and even the history of ML algorithm development. 
}

\emph{\textbf{Gap.} There is no systematic way to explore crypto-friendly alternative ML algorithms. The current practice is to design a problem/algorithm-specific crypto-friendly solution. Although the problem-specific design experiences and learnings can extend to a new solution design, there are no well-known rules or general frameworks for exploring such alternative ML algorithms yet.
}

\section{Security Proofs, Attacks, and Correctness}
In this section, we summarize the three aspects: security proofs, attack analysis, and correctness for existing CML approaches, which are commonly discussed in other cryptographic protocols.

\textbf{Security Proofs.} Homogeneous approaches do not use complex protocols other than the cryptographic primitive they use. For example, homomorphic encryption-based approaches involve only simple interactions between the client and the cloud - the client submitting the data and the cloud computes and returns the result; the GC-based methods have two involved parties following the fundamental GC protocols. Thus, most such approaches simply skip the security proof step, fully depending on the proven security and privacy guarantees provided by the primitives. 

For hybrid approaches, it's more sophisticated to prove their security, as they may include complex interactions among parties. We have observed two security proof frameworks are in prevalence. SecureML \cite{mohassel17} utilizes the Universally Composable Security (UC) framework \cite{canetti20}. The UC security framework defines security-preserving universal composition operation and allows for modular design and analysis of complex cryptographic protocols from simpler building blocks.  PrivateGraph \cite{sharma18tkde}, SecureBoost \cite{sharma19}, and Lu et al. \cite{lu17} adopt the simulation-based security proof \cite{lindell17}.  The simulation approach needs to show the existence of a \emph{simulator} in the ideal scenario that corresponds to the adversary in the real scenario, such that it is impossible to distinguish the interactions in the ideal scenario from those in the real scenario. The assumption of semi-honest parties held by most CML approaches makes the security proofs much easier \cite{lindell17,canetti20}.  As a result, many CML approaches ignore the steps of security proof. 

\textbf{Attacks.} To our knowledge, attacks on the confidentiality of cryptographic CML approaches have not been fully explored. Most works we covered in this category did not mention any potential attacks on their approaches, partially due to the well-known security guarantees provided by the underlying primitives or formal security proofs provided by a few approaches. While all approaches want to fully protect feature vectors in the training data, some approaches require the labels (in supervised learning) to be exposed for easier modeling \cite{graepel12}, and some even expose the final learned models \cite{niko13sp,lu17}. However, recent studies have shown that exposed models may lead to serious attacks, such as model inversion attacks \cite{fredrikson15,tramer16}, and membership inference attacks \cite{shokri16}.  

\textbf{Correctness.} Contrary to some cryptographic protocols and encryption systems that need to prove their correctness (e.g., encrypted values can be correctly decrypted), the correctness of CML protocols is attached to the correctness of the original machine learning algorithms. The DMC procedure honestly reassembles the original learning algorithm with the cryptographic components. Thus, as long as the primitives preserve the correctness and the composition strategy does not change the correctness (see Section \ref{sec:composition}), the correctness property is guaranteed. However, when researchers adopt a crypto-friendly alternative algorithm or component, they must justify whether the alternative methods warrant/attain the desired learning objective. SecureBoost \cite{sharma19} depends on the basic boosting theory \cite{schapire99} that states any weak base classifier, including random weak linear classifiers, can be used for the boosting framework. Naehrig et al. \cite{naehrig11} utilize the Taylor approximation of exponentiation to approximate the sigmoid function, which is a well-accepted mathematical method. While these alternative methods may affect the model quality, implying a potential trade-off between model quality and costs, they are all considered correct algorithms.

\emph{\textbf{Gap.} Security proofs are missing for some existing CML approaches, which raise a concern that they may contain flaws leading to significant information leaks. Further studies are needed to rigorously analyze these approaches.
}

\section{Evaluation Methods}
Researchers evaluate their proposed CML methods primarily based on costs and model quality. Some CML methods also involve trade-offs between these two aspects. 

\textbf{Costs.} CML researchers primarily concern about the costs of protocol, striving to find the most efficient secure protocols. Since multiple parties are involved, the costs for each party, i.e., the cloud provider, the client, and possibly the crypto-service provider or the second cloud provider, are all essential to the design of CML protocols. For a given CML method, each party's costs are the outcome of the cost for comparing the encryption/
decryption, data transmission, and other computation overhead. Because of the original motivation of outsourcing large-scale
computation, a skewed cost distribution between the client and the cloud is fundamental, i.e., the client should take much lower overheads compared to the cloud \cite{sharma18tkde,sharma19,mohassel17}. However, the client may still take much higher costs when running CML protocols when compared to running the original non-secure ML solution. The cost of external storage and related I/O operations are also critical to the cloud-side components as they are responsible for storing the encrypted data, which often is much larger than the plaintext version and cannot reside in memory. It is also highly desired that the cloud-side computation can be done parallelly with a popular processing framework such as MapReduce \cite{dean04,sharma18tkde}. Besides, when GC is adopted as a primitive to implement some components, additional communication cost related to the GC protocol is also significant, including the cost of transmitting the circuit and one-party's input data obliviously to the other party \cite{liu15,huang11}. As a result, the use of GC is limited to a few operations, such as comparison \cite{demmler15}. The overall computation and communication costs of different approaches are frequently compared and used as a measure to show the novelty of a new method. For example, Mohassel et al. \cite{mohassel17} show their work is more computation efficient than the GC-based framework considered by \cite{niko13sp} by about two orders of magnitude. Similarly, Sharma et al. \cite{sharma19} show their boosting solution is about three times faster than the neural network CML in \cite{mohassel17}. 

\textbf{Model Quality.} Model quality, a unique feature of CML evaluation,  is often tightly related to the cost of model training. Many machine learning algorithms are iterative, such as logistic regression, neural networks, and many clustering algorithms. As a result, model quality increases with the number of iterations until the process converges. However, a large number of iterations implies the increased overall costs. Some CML methods, e.g., Lu et al. \cite{lu17}, may only report the overall costs for one/few iterations of a specific learning algorithm, which is insufficient unless the number of iterations necessary for optimal results is specified. More precisely, many works miss the requirement that model evaluation should be tied to the cost evaluation, i.e., how much cost is needed to reach a certain model accuracy \cite{mohassel17,sharma19}. The discussion on crypto-friendly alternative algorithms also holds the assumption that model quality can be possibly traded off with costs, with the expectation that the crypto-friendly alternative may perform comparably or slightly worse than the original machine learning algorithm \cite{sharma19,sharma18tkde,mohassel17,graepel12}. 

\section{Other CML Approaches} \label{sec:other}
So far, we have focused on cryptographic methods based on well-known primitives. To cover a panoramic view of development in the growing area of confidential machine learning, we briefly discuss two closely related approaches, the perturbation-based approach and the hardware-assisted approach.

\subsection{Perturbation Methods} 
Most practical CML solutions that carefully follow the DMC process with some innovative uses of crypto-friendly ML algorithms still cost magnitudes more than the original plaintext algorithms. Especially if the learning algorithm is intrinsically expensive or relies on a massive-scale training dataset, the cryptographic primitives that provide semantic security may become impractically expensive, discouraging users from adopting the outsourcing paradigm. Another category of work: the perturbation-based approach offers much more efficient solutions with some weaker security notions. Often, they do not guarantee semantic security and may only be resilient to ciphertext-only attacks. Nevertheless, they can be interesting for users who are willing to make a practical trade-off between efficiency and the level of protection. We briefly discuss this body of work to extend readers' interests to this unique domain. 

The basic idea of perturbation is injecting random noises into the outsourced data while (approximately) preserving some specific properties machine learning models rely upon. The most well-known properties are geometric and topological structures in the multidimensional space. Therefore, one can still train a model from the perturbed data on the untrusted platform with preserved confidentiality of both data and model. Typical perturbation methods include randomized response \cite{erlingsson14,du03kdd}, additive perturbation \cite{rakesh00}, geometric perturbation \cite{keke11kais}, random projection perturbation \cite{liu06tkde}, and random space perturbation \cite{xu14}. They have been applied to decision tree learning \cite{du03kdd,rakesh00}, clustering \cite{keke11kais,liu06tkde}, kNN classifier \cite{keke11kais}, support vector machines \cite{keke11kais}, linear classifier \cite{keke11kais,chen18}, and boosting \cite{chen18}. The perturbation mechanisms can also disguise the training images in deep learning frameworks \cite{sharma18dl} to achieve much lower training costs than cryptographic protocols \cite{mohassel17}. Furthermore, the perturbation methods often do not involve expensive cryptographic primitives. Consequentially, one can observe significant cost savings in the entire life cycle of data analytics, including data submission, computation, and communication amongst the involved parties.

\emph{\textbf{Insight.} The key idea of perturbation approaches is to identify a certain high-level utility and preserve it in secure randomized transformations. Similar ideas have also been explored in the cryptographic domain, such as order-preserving encryption \cite{boldyreva11,boldyreva09,kerschbaum15} and encrypted keyword search \cite{golle04,reza06}}.

\emph{\textbf{Gap.}
Despite their efficiency, perturbation approaches face two critical weaknesses. First, perturbation methods may cause significant degradation to the data quality and introduce significant trade-offs between utility and confidentiality. Second, there is no systematic framework for analyzing the protection level guaranteed by a perturbation method.   Some of them are known not to provide provable semantic security \cite{keke11kais,xu14}. However, under a clear, rigorous threat model definition and thorough analysis, these methods will have high practical values in the venues where users can accept the specific threat model.   
}
\subsection{Hardware-Assisted Approaches}
During the past few years, hardware-assisted trustworthy computing has made a significant breakthrough. In particular, several CPU manufactures have implemented the trusted execution environment (TEE) platforms, among which the most popular one is Intel's Software Guard Extensions (SGX) \cite{sgx-explained}. We will take SGX as an example in the following. SGX defines a specific memory area (e.g., the \emph{enclave}). Only the authorized owner can run programs and access data in the enclave via special instructions. Owners and users gain access rights via an \emph{attestation} protocol. SGX minimized the trust boundary to the enclave, which means even though the entire operating system is compromised, adversaries cannot access the enclave. The physical enclave memory is limited (less than 100MB are usable by users). When the enclave memory pages are swapped out/in by the virtual memory management subsystem of the OS\footnote{The enclave virtual memory management is only enabled on the Linux system for early versions of SGX, which might be changed in newer versions of SGX}, they are encrypted/decrypted by the SGX library functions implicitly. SGX uses AES encryption (is this always true?), and thus the encryption and decryption costs are much lower than the primitives we have discussed so far. Besides, since the enclave program works on decrypted data, there is no need to develop special CML algorithms for running inside the enclave, making SGX an appealing platform for developing CML solutions for complex algorithms working with large data. 

However, there are a few challenges for migrating algorithms to the SGX environment. First, users need to learn the whole SGX working mechanism and learn to use special instructions and APIs, which can be inconvenient. A few efforts have simplified the migration of applications to SGX, among which the Graphene-SGX library OS \cite{tsai17}, SCONE \cite{arnautov16}, and Panoply \cite{shinde17} are the most well-known. With a tool like Graphene-SGX, developing CML solutions becomes more straightforward. Lee et al. \cite{lee20} have tried to migrate machine learning algorithms to SGX based on Graphene-SGX. However, these methods do not address side-channel attacks.

Second, side-channel attacks are considered the primary threat to SGX-based applications. As TEEs have prevented many traditional attacks and the assumption is now changed to adversary-controlled OS, side-channel attacks are active research areas. Memory side channels and cache side channels are the two types that researchers mostly examined. Memory side-channel attacks are primarily access pattern attacks \cite{sasy18,ahmad18,shinde16}. As the encrypted data have to be loaded from the file to the untrusted area first and then accessed by the enclave, the access pattern attacks seem inevitable for data-intensive applications like CML. The well-known approach addressing this problem is the Oblivious RAM technique \cite{oded96}, which has been applied to SGX by ZeroTrace \cite{sasy18} and Obliviate \cite{ahmad18}. Ohrimenko et al. \cite{ohrimenko16} also used oblivious access techniques for multi-party machine learning with SGX. Branching attacks \cite{shinde16} utilize the branching statements and manipulate page faults to extract information, often addressable with oblivious branching instructions such as CMOV \cite{shinde16,sasy18,alam21}. Cache side-channel attacks such as cache timing and transient execution state \cite{bulck18,ristenpart09,kocher19,lipp18} utilize the unique CPU architectural features and thus depend on the manufacturers' firmware and software patches to fix. More studies are necessary to explore the full potential and unique problems with SGX-based CML.

\emph{\textbf{Insight.} The TEE, e.g., SGX, techniques can significantly boost CML's performance on untrusted platforms, as the solutions do not involve expensive crypto primitives or protocols. We consider the SGX based CML as a promising direction because it achieves a strong confidentiality guarantee with significant performance benefits compared to other approaches.  
}

\emph{\textbf{Gap.} The most critical challenge TEEs face is side-channel attacks, especially the access pattern attacks. Also, machine learning algorithms have unique features (e.g., data access, batching, etc.) that may lead to specific attacks that have not been fully explored yet. Another practical concern is that most recent Intel server CPUs still have not had SGX enabled. A few cloud platforms such as Microsoft Azure and IBM Cloud have started offering SGX-enabled instances, and thus we consider this gap of missing public SGX resources will be filled up soon.}

\section{Conclusion}\label{sec:conclusion}
Despite the potential risk of data and model leakages, many resource-constrained data owners use untrusted platforms (e.g., clouds and edges) for training machine learning models. Researchers have been designing and developing confidential machine learning (CML) approaches for outsourced data using cryptographic primitives and various composition strategies. CML's overall goal is to protect the confidentiality of data, model, and intermediate results from the untrusted platforms while also preserving the trained model quality with acceptable costs. 

We have reviewed the recent significant developments on CML under a systemization framework, focusing on the cryptographic approaches. We have included the cryptographic primitives that are the backbone of the CML approaches and compared the costs for basic operations. While the homogeneous methods that rely on a single cryptographic primitive are straightforward, their solutions are too expensive to be practical. Thus, we focus on the primary design trend of the hybrid composition of multiple primitives under the decomposition-mapping-composition (DMC) procedure and the selection of crypto-friendly alternative learning algorithms. We describe the critical issues such as the switching between primitives and the principles of identifying crypto-friendly machine learning algorithms. Finally, we also include a brief discussion of related approaches and new directions, including the perturbation and hardware-assisted methods. At the end of most sections, we have also included a concise summary area labeled with \emph{Insight} and \emph{Gap} for readers to get the gist conveniently. We believe this survey can be valuable to both researchers and practitioners to build more complex and practical CML solutions in the future.

\begin{backmatter}

\section*{Acknowledgements}
Not applicable

\section*{Funding}
This work is partially supported by the National Science Foundation under grant no. 1245847 and the National Institute of Health under grant no. 1R43AI136357-01A1.  


\section*{Availability of data and materials}
Not applicable








\bibliographystyle{bmc-mathphys} 

\newcommand{\BMCxmlcomment}[1]{}

\BMCxmlcomment{

<refgrp>

<bibl id="B1">
  <title><p>Toward Practical Privacy-Preserving Analytics for IoT and
  Cloud-Based Healthcare Systems</p></title>
  <aug>
    <au><snm>Sharma</snm><fnm>S</fnm></au>
    <au><snm>Chen</snm><fnm>K</fnm></au>
    <au><snm>Sheth</snm><fnm>A</fnm></au>
  </aug>
  <source>IEEE Internet Computing</source>
  <pubdate>2018</pubdate>
  <volume>22</volume>
  <issue>2</issue>
  <fpage>42</fpage>
  <lpage>51</lpage>
  <url>doi.ieeecomputersociety.org/10.1109/MIC.2018.112102519</url>
</bibl>

<bibl id="B2">
  <title><p>Insider Attacks in Cloud Computing</p></title>
  <aug>
    <au><snm>Duncan</snm><fnm>A. J.</fnm></au>
    <au><snm>Creese</snm><fnm>S.</fnm></au>
    <au><snm>Goldsmith</snm><fnm>M.</fnm></au>
  </aug>
  <source>2012 IEEE 11th International Conference on Trust, Security and
  Privacy in Computing and Communications</source>
  <pubdate>2012</pubdate>
</bibl>

<bibl id="B3">
  <title><p>GCreep: Google engineer stalked teens, spied on chats.</p></title>
  <aug>
    <au><snm>Chen</snm><fnm>A</fnm></au>
  </aug>
  <source>Gawker, http://gawker.com/5637234/</source>
  <pubdate>2010</pubdate>
</bibl>

<bibl id="B4">
  <title><p>The {A}shley {M}adison affair.</p></title>
  <aug>
    <au><snm>Mansfield Devine</snm><fnm>S</fnm></au>
  </aug>
  <source>Network Security</source>
  <pubdate>2015</pubdate>
  <volume>2015</volume>
  <issue>9</issue>
  <fpage>8</fpage>
  <lpage>16</lpage>
  <url>http://ezproxy.libraries.wright.edu/login?url=http://search.ebscohost.com/login.aspx?direct=true&db=bth&AN=109503050&site=eds-live</url>
</bibl>

<bibl id="B5">
  <title><p>Breaches to Customer Account Data.</p></title>
  <aug>
    <au><snm>Unger</snm><fnm>L</fnm></au>
  </aug>
  <source>Computer and Internet Lawyer</source>
  <pubdate>2015</pubdate>
  <volume>32</volume>
  <issue>2</issue>
  <fpage>14</fpage>
  <lpage>20</lpage>
  <url>http://ezproxy.libraries.wright.edu/login?url=http://search.ebscohost.com/login.aspx?direct=true&db=cph&AN=100337545&site=eds-live</url>
</bibl>

<bibl id="B6">
  <title><p>Fully homomorphic encryption using ideal lattices</p></title>
  <aug>
    <au><snm>Gentry</snm><fnm>C</fnm></au>
  </aug>
  <source>{Annual ACM Symposium on Theory of Computing}</source>
  <publisher>New York, NY, USA: ACM</publisher>
  <pubdate>2009</pubdate>
  <fpage>169</fpage>
  <lpage>-178</lpage>
</bibl>

<bibl id="B7">
  <title><p>How to generate and exhange secrets</p></title>
  <aug>
    <au><snm>Yao</snm><fnm>AC</fnm></au>
  </aug>
  <source>IEEE Symposium on Foundations of Computer Science</source>
  <pubdate>1986</pubdate>
  <fpage>162</fpage>
  <lpage>167</lpage>
</bibl>

<bibl id="B8">
  <title><p>Privacy-preserving matrix factorization</p></title>
  <aug>
    <au><snm>Nikolaenko</snm><fnm>V</fnm></au>
    <au><snm>Ioannidis</snm><fnm>S</fnm></au>
    <au><snm>Weinsberg</snm><fnm>U</fnm></au>
    <au><snm>Joye</snm><fnm>M</fnm></au>
    <au><snm>Taft</snm><fnm>N</fnm></au>
    <au><snm>Boneh</snm><fnm>D</fnm></au>
  </aug>
  <source>ACM SIGSAC conference on Computer and communications
  security</source>
  <pubdate>2013</pubdate>
  <fpage>801</fpage>
  <lpage>-812</lpage>
</bibl>

<bibl id="B9">
  <title><p>Privacy-Preserving Ridge Regression on Hundreds of Millions of
  Records</p></title>
  <aug>
    <au><snm>Nikolaenko</snm><fnm>V</fnm></au>
    <au><snm>Weinsberg</snm><fnm>U</fnm></au>
    <au><snm>Ioannidis</snm><fnm>S</fnm></au>
    <au><snm>Joye</snm><fnm>M</fnm></au>
    <au><snm>Boneh</snm><fnm>D</fnm></au>
    <au><snm>Taft</snm><fnm>N</fnm></au>
  </aug>
  <source>IEEE Symposium on Security and Privacy</source>
  <pubdate>2013</pubdate>
  <fpage>334</fpage>
  <lpage>-348</lpage>
</bibl>

<bibl id="B10">
  <title><p>{ABY} - {A} Framework for Efficient Mixed-Protocol Secure Two-Party
  Computation</p></title>
  <aug>
    <au><snm>Demmler</snm><fnm>D</fnm></au>
    <au><snm>Schneider</snm><fnm>T</fnm></au>
    <au><snm>Zohner</snm><fnm>M</fnm></au>
  </aug>
  <source>22nd Annual Network and Distributed System Security Symposium, {NDSS}
  2015, San Diego, California, USA, February 8-11, 2015</source>
  <pubdate>2015</pubdate>
  <url>https://www.ndss-symposium.org/ndss2015/aby---framework-efficient-mixed-protocol-secure-two-party-computation</url>
</bibl>

<bibl id="B11">
  <title><p>SecureML: A System for Scalable Privacy-Preserving Machine
  Learning</p></title>
  <aug>
    <au><snm>Mohassel</snm><fnm>P.</fnm></au>
    <au><snm>Zhang</snm><fnm>Y.</fnm></au>
  </aug>
  <source>2017 IEEE Symposium on Security and Privacy (SP)</source>
  <pubdate>2017</pubdate>
  <fpage>19</fpage>
  <lpage>38</lpage>
</bibl>

<bibl id="B12">
  <title><p>PrivateGraph: Privacy-Preserving Spectral Analysis of Encrypted
  Graphs in the Cloud</p></title>
  <aug>
    <au><snm>Sharma</snm><fnm>S</fnm></au>
    <au><snm>Powers</snm><fnm>J</fnm></au>
    <au><snm>Chen</snm><fnm>K</fnm></au>
  </aug>
  <source>IEEE Transactions on Knowledge and Data Engineering</source>
  <pubdate>2019</pubdate>
  <volume>31</volume>
  <issue>5</issue>
  <fpage>981</fpage>
  <lpage>995</lpage>
</bibl>

<bibl id="B13">
  <title><p>Confidential Boosting with Random Linear Classifiers for Outsourced
  User-Generated Data</p></title>
  <aug>
    <au><snm>Sharma</snm><fnm>S</fnm></au>
    <au><snm>Chen</snm><fnm>K</fnm></au>
  </aug>
  <source>Computer Security - {ESORICS} 2019 - 24th European Symposium on
  Research in Computer Security, Luxembourg, September 23-27, 2019,
  Proceedings, Part {I}</source>
  <pubdate>2019</pubdate>
  <fpage>41</fpage>
  <lpage>-65</lpage>
</bibl>

<bibl id="B14">
  <title><p>CryptoNets: Applying Neural Networks to Encrypted Data with High
  Throughput and Accuracy</p></title>
  <aug>
    <au><snm>Gilad Bachrach</snm><fnm>R</fnm></au>
    <au><snm>Dowlin</snm><fnm>N</fnm></au>
    <au><snm>Laine</snm><fnm>K</fnm></au>
    <au><snm>Lauter</snm><fnm>K</fnm></au>
    <au><snm>Naehrig</snm><fnm>M</fnm></au>
    <au><snm>Wernsing</snm><fnm>J</fnm></au>
  </aug>
  <source>Proceedings of The 33rd International Conference on Machine
  Learning</source>
  <editor>Maria Florina Balcan and Kilian Q. Weinberger</editor>
  <series><title><p>Proceedings of Machine Learning
  Research</p></title></series>
  <pubdate>2016</pubdate>
  <volume>48</volume>
  <fpage>201</fpage>
  <lpage>-210</lpage>
</bibl>

<bibl id="B15">
  <title><p>Machine Learning Classification over Encrypted Data</p></title>
  <aug>
    <au><snm>Bost</snm><fnm>R</fnm></au>
    <au><snm>Popa</snm><fnm>RA</fnm></au>
    <au><snm>Tu</snm><fnm>S</fnm></au>
    <au><snm>Goldwasser</snm><fnm>S</fnm></au>
  </aug>
  <source>Annual Network and Distributed System Security Symposium
  (NDSS)</source>
  <pubdate>2015</pubdate>
</bibl>

<bibl id="B16">
  <title><p>CryptoDL: Deep Neural Networks over Encrypted Data</p></title>
  <aug>
    <au><snm>Hesamifard</snm><fnm>E</fnm></au>
    <au><snm>Takabi</snm><fnm>H</fnm></au>
    <au><snm>Ghasemi</snm><fnm>M</fnm></au>
  </aug>
  <source>CoRR</source>
  <pubdate>2017</pubdate>
  <volume>abs/1711.05189</volume>
  <url>http://arxiv.org/abs/1711.05189</url>
</bibl>

<bibl id="B17">
  <title><p>ReDCrypt: RealTime Privacy Preserving Deep Learning Using
  FPGAs</p></title>
  <aug>
    <au><snm>Rouhani</snm><fnm>B</fnm></au>
    <au><snm>Hussain</snm><fnm>SU</fnm></au>
    <au><snm>Lauter</snm><fnm>K</fnm></au>
    <au><snm>Koushanfar</snm><fnm>F</fnm></au>
  </aug>
  <source>ACM Transactions on Reconfigurable Technology and Systems
  (TRETS)</source>
  <pubdate>2018</pubdate>
</bibl>

<bibl id="B18">
  <title><p>The Elements of Statistical Learning</p></title>
  <aug>
    <au><snm>Hastie</snm><fnm>T</fnm></au>
    <au><snm>Tibshirani</snm><fnm>R</fnm></au>
    <au><snm>Friedman</snm><fnm>J</fnm></au>
  </aug>
  <publisher>New York City, New York: Springer-Verlag</publisher>
  <pubdate>2001</pubdate>
</bibl>

<bibl id="B19">
  <title><p>Practical Secure Computation Outsourcing: A Survey</p></title>
  <aug>
    <au><snm>Shan</snm><fnm>Z</fnm></au>
    <au><snm>Ren</snm><fnm>K</fnm></au>
    <au><snm>Blanton</snm><fnm>M</fnm></au>
    <au><snm>Wang</snm><fnm>C</fnm></au>
  </aug>
  <source>ACM COMPUTING SURVEYS</source>
  <pubdate>2018</pubdate>
  <volume>51</volume>
  <issue>2</issue>
</bibl>

<bibl id="B20">
  <title><p>A survey of deep learning techniques for autonomous
  driving</p></title>
  <aug>
    <au><snm>Grigorescu</snm><fnm>S</fnm></au>
    <au><snm>Trasnea</snm><fnm>B</fnm></au>
    <au><snm>Cocias</snm><fnm>T</fnm></au>
    <au><snm>Macesanu</snm><fnm>G</fnm></au>
  </aug>
  <source>Journal of Field Robotics</source>
  <pubdate>2019</pubdate>
  <volume>37</volume>
  <issue>3</issue>
</bibl>

<bibl id="B21">
  <title><p>SoK: Security and Privacy in Machine Learning</p></title>
  <aug>
    <au><snm>{Papernot}</snm><fnm>N.</fnm></au>
    <au><snm>{McDaniel}</snm><fnm>P.</fnm></au>
    <au><snm>{Sinha}</snm><fnm>A.</fnm></au>
    <au><snm>{Wellman}</snm><fnm>M. P.</fnm></au>
  </aug>
  <source>2018 IEEE European Symposium on Security and Privacy (EuroS
  P)</source>
  <pubdate>2018</pubdate>
  <fpage>399</fpage>
  <lpage>414</lpage>
</bibl>

<bibl id="B22">
  <title><p>A Survey on Homomorphic Encryption Schemes: Theory and
  Implementation</p></title>
  <aug>
    <au><snm>Acar</snm><fnm>A</fnm></au>
    <au><snm>Aksu</snm><fnm>H</fnm></au>
    <au><snm>Uluagac</snm><fnm>AS</fnm></au>
    <au><snm>Conti</snm><fnm>M</fnm></au>
  </aug>
  <source>ACM Computing Surveys</source>
  <publisher>New York, NY, USA: Association for Computing Machinery</publisher>
  <pubdate>2018</pubdate>
  <volume>51</volume>
  <issue>4</issue>
  <url>https://doi.org/10.1145/3214303</url>
</bibl>

<bibl id="B23">
  <title><p>Secure Multiparty Computation (MPC)</p></title>
  <aug>
    <au><snm>Lindell</snm><fnm>Y</fnm></au>
  </aug>
  <source>Cryptology ePrint Archive, Report 2020/300</source>
  <pubdate>2020</pubdate>
  <note>\url{https://eprint.iacr.org/2020/300}</note>
</bibl>

<bibl id="B24">
  <aug>
    <au><snm>{Evans}</snm><fnm>D.</fnm></au>
    <au><snm>{Kolesnikov}</snm><fnm>V.</fnm></au>
    <au><snm>{Rosulek}</snm><fnm>M.</fnm></au>
  </aug>
  <source>A Pragmatic Introduction to Secure Multi-Party Computation</source>
  <pubdate>2018</pubdate>
</bibl>

<bibl id="B25">
  <title><p>Deep Learning with Differential Privacy</p></title>
  <aug>
    <au><snm>Abadi</snm><fnm>M</fnm></au>
    <au><snm>Chu</snm><fnm>A</fnm></au>
    <au><snm>Goodfellow</snm><fnm>I</fnm></au>
    <au><snm>McMahan</snm><fnm>HB</fnm></au>
    <au><snm>Mironov</snm><fnm>I</fnm></au>
    <au><snm>Talwar</snm><fnm>K</fnm></au>
    <au><snm>Zhang</snm><fnm>L</fnm></au>
  </aug>
  <source>Proceedings of the 2016 ACM SIGSAC Conference on Computer and
  Communications Security</source>
  <publisher>New York, NY, USA: ACM</publisher>
  <series><title><p>CCS '16</p></title></series>
  <pubdate>2016</pubdate>
  <fpage>308</fpage>
  <lpage>-318</lpage>
  <url>http://doi.acm.org/10.1145/2976749.2978318</url>
</bibl>

<bibl id="B26">
  <title><p>Privacy-preserving deep learning</p></title>
  <aug>
    <au><snm>Shokri</snm><fnm>R</fnm></au>
    <au><snm>Shmatikov</snm><fnm>V</fnm></au>
  </aug>
  <source>Proceedings of the 22nd ACM SIGSAC Conference on Computer and
  Communications Security</source>
  <pubdate>2015</pubdate>
</bibl>

<bibl id="B27">
  <title><p>Differential Privacy and Machine Learning: a Survey and
  Review</p></title>
  <aug>
    <au><snm>Ji</snm><fnm>Z</fnm></au>
    <au><snm>Lipton</snm><fnm>ZC</fnm></au>
    <au><snm>Elkan</snm><fnm>C</fnm></au>
  </aug>
  <source>CoRR</source>
  <pubdate>2014</pubdate>
  <volume>abs/1412.7584</volume>
  <url>http://arxiv.org/abs/1412.7584</url>
</bibl>

<bibl id="B28">
  <title><p>Signal Processing and Machine Learning with Differential Privacy:
  Algorithms and Challenges for Continuous Data</p></title>
  <aug>
    <au><snm>Sarwate</snm><fnm>A. D.</fnm></au>
    <au><snm>Chaudhuri</snm><fnm>K.</fnm></au>
  </aug>
  <source>IEEE Signal Processing Magazine</source>
  <pubdate>2013</pubdate>
  <volume>30</volume>
  <issue>5</issue>
  <fpage>86</fpage>
  <lpage>94</lpage>
</bibl>

<bibl id="B29">
  <title><p>Privacy-Preserving Data Mining: Models and Algorithms</p></title>
  <aug>
    <au><snm>Aggarwal</snm><fnm>CC</fnm></au>
    <au><snm>Yu</snm><fnm>PS</fnm></au>
  </aug>
  <publisher>New York City, NY: Springer</publisher>
  <pubdate>2010</pubdate>
</bibl>

<bibl id="B30">
  <title><p>Privacy-Preserving Data Mining Techniques: Survey and
  Challenges</p></title>
  <aug>
    <au><snm>Matwin</snm><fnm>S</fnm></au>
  </aug>
  <source>Discrimination and Privacy in the Information Society: Data Mining
  and Profiling in Large Databases</source>
  <publisher>Berlin, Heidelberg: Springer Berlin Heidelberg</publisher>
  <editor>Custers, Bart and Calders, Toon and Schermer, Bart and Zarsky,
  Tal</editor>
  <pubdate>2013</pubdate>
  <fpage>209</fpage>
  <lpage>-221</lpage>
</bibl>

<bibl id="B31">
  <title><p>A comprehensive review on privacy preserving data
  mining</p></title>
  <aug>
    <au><snm>Aldeen</snm><fnm>YAAS</fnm></au>
    <au><snm>Salleh</snm><fnm>M</fnm></au>
    <au><snm>Razzaque</snm><fnm>MA</fnm></au>
  </aug>
  <source>SpringerPlus</source>
  <pubdate>2015</pubdate>
  <volume>4</volume>
  <issue>1</issue>
  <fpage>694</fpage>
  <url>https://doi.org/10.1186/s40064-015-1481-x</url>
</bibl>

<bibl id="B32">
  <title><p>An Analysis of Privacy Preservation Techniques in Data
  Mining</p></title>
  <aug>
    <au><snm>Sachan</snm><fnm>A</fnm></au>
    <au><snm>Roy</snm><fnm>D</fnm></au>
    <au><snm>Arun</snm><fnm>P. V.</fnm></au>
  </aug>
  <source>Advances in Computing and Information Technology</source>
  <publisher>Berlin, Heidelberg: Springer Berlin Heidelberg</publisher>
  <editor>Meghanathan, Natarajan and Nagamalai, Dhinaharan and Chaki,
  Nabendu</editor>
  <pubdate>2013</pubdate>
  <fpage>119</fpage>
  <lpage>-128</lpage>
</bibl>

<bibl id="B33">
  <title><p>ML Confidential: Machine Learning on Encrypted Data</p></title>
  <aug>
    <au><snm>Graepel</snm><fnm>T</fnm></au>
    <au><snm>Lauter</snm><fnm>K</fnm></au>
    <au><snm>Naehrig</snm><fnm>M</fnm></au>
  </aug>
  <source>International Conference on Information Security and
  Cryptology</source>
  <pubdate>2013</pubdate>
  <fpage>1</fpage>
  <lpage>-21</lpage>
</bibl>

<bibl id="B34">
  <title><p>Using Fully Homomorphic Encryption for Statistical Analysis of
  Categorical, Ordinal and Numerical Data</p></title>
  <aug>
    <au><snm>Lu</snm><fnm>W</fnm></au>
    <au><snm>Kawasaki</snm><fnm>S</fnm></au>
    <au><snm>Sakuma</snm><fnm>J</fnm></au>
  </aug>
  <source>The Network and Distributed System Security Symposium</source>
  <pubdate>2017</pubdate>
</bibl>

<bibl id="B35">
  <title><p>A Survey on Security Threats and Defensive Techniques of Machine
  Learning: A Data Driven View</p></title>
  <aug>
    <au><snm>{Liu}</snm><fnm>Q.</fnm></au>
    <au><snm>{Li}</snm><fnm>P.</fnm></au>
    <au><snm>{Zhao}</snm><fnm>W.</fnm></au>
    <au><snm>{Cai}</snm><fnm>W.</fnm></au>
    <au><snm>{Yu}</snm><fnm>S.</fnm></au>
    <au><snm>{Leung}</snm><fnm>V. C. M.</fnm></au>
  </aug>
  <source>IEEE Access</source>
  <pubdate>2018</pubdate>
  <volume>6</volume>
  <fpage>12103</fpage>
  <lpage>12117</lpage>
</bibl>

<bibl id="B36">
  <title><p>Privacy in Pharmacogenetics: An End-to-End Case Study of
  Personalized Warfarin Dosing</p></title>
  <aug>
    <au><snm>Fredrikson</snm><fnm>M</fnm></au>
    <au><snm>Lantz</snm><fnm>E</fnm></au>
    <au><snm>Jha</snm><fnm>S</fnm></au>
    <au><snm>Lin</snm><fnm>S</fnm></au>
    <au><snm>Page</snm><fnm>D</fnm></au>
    <au><snm>Ristenpart</snm><fnm>T</fnm></au>
  </aug>
  <source>23rd USENIX Security Symposium USENIX Security</source>
  <publisher>San Diego, CA: USENIX Association</publisher>
  <pubdate>2014</pubdate>
  <fpage>17</fpage>
  <lpage>-32</lpage>
</bibl>

<bibl id="B37">
  <title><p>Model Inversion Attacks that Exploit Confidence Information and
  Basic Countermeasures</p></title>
  <aug>
    <au><snm>Fredrikson</snm><fnm>M</fnm></au>
    <au><snm>Jha</snm><fnm>S</fnm></au>
    <au><snm>Ristenpart</snm><fnm>T</fnm></au>
  </aug>
  <source>ACM Conference on Computer and Communications Security</source>
  <pubdate>2015</pubdate>
</bibl>

<bibl id="B38">
  <title><p>Deep Models Under the GAN: Information Leakage from Collaborative
  Deep Learning</p></title>
  <aug>
    <au><snm>Hitaj</snm><fnm>B</fnm></au>
    <au><snm>Ateniese</snm><fnm>G</fnm></au>
    <au><snm>Perez Cruz</snm><fnm>F</fnm></au>
  </aug>
  <source>Proceedings of the 2017 ACM SIGSAC Conference on Computer and
  Communications Security</source>
  <publisher>New York, NY, USA: ACM</publisher>
  <series><title><p>CCS '17</p></title></series>
  <pubdate>2017</pubdate>
  <fpage>603</fpage>
  <lpage>-618</lpage>
  <url>http://doi.acm.org/10.1145/3133956.3134012</url>
</bibl>

<bibl id="B39">
  <title><p>Membership Inference Attacks Against Machine Learning
  Models</p></title>
  <aug>
    <au><snm>Shokri</snm><fnm>R</fnm></au>
    <au><snm>Stronati</snm><fnm>M</fnm></au>
    <au><snm>Song</snm><fnm>C</fnm></au>
    <au><snm>Shmatikov</snm><fnm>V</fnm></au>
  </aug>
  <source>2017 {IEEE} Symposium on Security and Privacy, {SP} 2017, San Jose,
  CA, USA, May 22-26, 2017</source>
  <pubdate>2017</pubdate>
  <fpage>3</fpage>
  <lpage>-18</lpage>
</bibl>

<bibl id="B40">
  <title><p>Overlearning Reveals Sensitive Attributes</p></title>
  <aug>
    <au><snm>Song</snm><fnm>C</fnm></au>
    <au><snm>Shmatikov</snm><fnm>V</fnm></au>
  </aug>
  <source>International Conference on Learning Representations</source>
  <pubdate>2020</pubdate>
  <url>https://openreview.net/forum?id=SJeNz04tDS</url>
</bibl>

<bibl id="B41">
  <title><p>Public-Key Cryptosystems Based on Composite Degree Residuosity
  Classes</p></title>
  <aug>
    <au><snm>Paillier</snm><fnm>P</fnm></au>
  </aug>
  <source>The proceedings of EUROCRYPT</source>
  <pubdate>1999</pubdate>
  <fpage>223</fpage>
  <lpage>-238</lpage>
</bibl>

<bibl id="B42">
  <title><p>{TFHE:} Fast Fully Homomorphic Encryption Over the
  Torus</p></title>
  <aug>
    <au><snm>Chillotti</snm><fnm>I</fnm></au>
    <au><snm>Gama</snm><fnm>N</fnm></au>
    <au><snm>Georgieva</snm><fnm>M</fnm></au>
    <au><snm>Izabach{\`{e}}ne</snm><fnm>M</fnm></au>
  </aug>
  <source>J. Cryptology</source>
  <pubdate>2020</pubdate>
  <volume>33</volume>
  <issue>1</issue>
  <fpage>34</fpage>
  <lpage>-91</lpage>
  <url>https://doi.org/10.1007/s00145-019-09319-x</url>
</bibl>

<bibl id="B43">
  <title><p>Homomorphic Encryption for Arithmetic of Approximate
  Numbers</p></title>
  <aug>
    <au><snm>Cheon</snm><fnm>JH</fnm></au>
    <au><snm>Kim</snm><fnm>A</fnm></au>
    <au><snm>Kim</snm><fnm>M</fnm></au>
    <au><snm>Song</snm><fnm>Y</fnm></au>
  </aug>
  <source>Advances in Cryptology -- ASIACRYPT 2017</source>
  <publisher>Cham: Springer International Publishing</publisher>
  <editor>Takagi, Tsuyoshi and Peyrin, Thomas</editor>
  <pubdate>2017</pubdate>
  <fpage>409</fpage>
  <lpage>-437</lpage>
</bibl>

<bibl id="B44">
  <title><p>(Leveled) Fully Homomorphic Encryption Without
  Bootstrapping</p></title>
  <aug>
    <au><snm>Brakerski</snm><fnm>Z</fnm></au>
    <au><snm>Gentry</snm><fnm>C</fnm></au>
    <au><snm>Vaikuntanathan</snm><fnm>V</fnm></au>
  </aug>
  <source>Innovations in Theoretical Computer Science Conference
  (ITSC)</source>
  <pubdate>2012</pubdate>
  <fpage>309</fpage>
  <lpage>-325</lpage>
</bibl>

<bibl id="B45">
  <title><p>Algorithms in {HEL}ib</p></title>
  <aug>
    <au><snm>Halevi</snm><fnm>S</fnm></au>
    <au><snm>Shoup</snm><fnm>V</fnm></au>
  </aug>
  <source>International Cryptology Conference</source>
  <pubdate>2014</pubdate>
  <fpage>554</fpage>
  <lpage>-571</lpage>
</bibl>

<bibl id="B46">
  <title><p>Improved Garbled Circuit: Free XOR Gates and
  Applications</p></title>
  <aug>
    <au><snm>Kolesnikov</snm><fnm>V</fnm></au>
    <au><snm>Schneider</snm><fnm>T</fnm></au>
  </aug>
  <source>Proceedings of the 35th International Colloquium on Automata,
  Languages and Programming, Part II</source>
  <publisher>Berlin, Heidelberg: Springer-Verlag</publisher>
  <pubdate>2008</pubdate>
  <fpage>486</fpage>
  <lpage>-498</lpage>
</bibl>

<bibl id="B47">
  <title><p>Two Halves Make a Whole</p></title>
  <aug>
    <au><snm>Zahur</snm><fnm>S</fnm></au>
    <au><snm>Rosulek</snm><fnm>M</fnm></au>
    <au><snm>Evans</snm><fnm>D</fnm></au>
  </aug>
  <source>Annual International Conference on the Theory and Applications of
  Cryptographic Techniques (EUROCRYPT)</source>
  <publisher>Berlin, Heidelberg: Springer Berlin Heidelberg</publisher>
  <pubdate>2015</pubdate>
  <fpage>220</fpage>
  <lpage>-250</lpage>
</bibl>

<bibl id="B48">
  <title><p>More efficient oblivious transfer and extensions for faster secure
  computation</p></title>
  <aug>
    <au><snm>Asharov</snm><fnm>G</fnm></au>
    <au><snm>Lindell</snm><fnm>Y</fnm></au>
    <au><snm>Schneider</snm><fnm>T</fnm></au>
    <au><snm>Zohner</snm><fnm>M</fnm></au>
  </aug>
  <source>2013 {ACM} {SIGSAC} Conference on Computer and Communications
  Security, CCS'13, Berlin, Germany, November 4-8, 2013</source>
  <pubdate>2013</pubdate>
  <fpage>535</fpage>
  <lpage>-548</lpage>
  <url>http://doi.acm.org/10.1145/2508859.2516738</url>
</bibl>

<bibl id="B49">
  <title><p>Faster Secure Two-party Computation Using Garbled
  Circuits</p></title>
  <aug>
    <au><snm>Huang</snm><fnm>Y</fnm></au>
    <au><snm>Evans</snm><fnm>D</fnm></au>
    <au><snm>Katz</snm><fnm>J</fnm></au>
    <au><snm>Malka</snm><fnm>L</fnm></au>
  </aug>
  <source>USENIX Conference on Security</source>
  <pubdate>2011</pubdate>
  <fpage>35</fpage>
  <lpage>-35</lpage>
</bibl>

<bibl id="B50">
  <title><p>ObliVM: A Programming Framework for Secure Computation</p></title>
  <aug>
    <au><snm>Liu</snm><fnm>C.</fnm></au>
    <au><snm>Wang</snm><fnm>X. S.</fnm></au>
    <au><snm>Nayak</snm><fnm>K.</fnm></au>
    <au><snm>Huang</snm><fnm>Y.</fnm></au>
    <au><snm>Shi</snm><fnm>E.</fnm></au>
  </aug>
  <source>2015 IEEE Symposium on Security and Privacy</source>
  <pubdate>2015</pubdate>
  <fpage>359</fpage>
  <lpage>376</lpage>
</bibl>

<bibl id="B51">
  <title><p>Top 10 Algorithms in Data Mining</p></title>
  <aug>
    <au><snm>Wu</snm><fnm>X</fnm></au>
    <au><snm>Kumar</snm><fnm>V</fnm></au>
    <au><snm>Ross Quinlan</snm><fnm>J.</fnm></au>
    <au><snm>Ghosh</snm><fnm>J</fnm></au>
    <au><snm>Yang</snm><fnm>Q</fnm></au>
    <au><snm>Motoda</snm><fnm>H</fnm></au>
    <au><snm>McLachlan</snm><fnm>GJ</fnm></au>
    <au><snm>Ng</snm><fnm>A</fnm></au>
    <au><snm>Liu</snm><fnm>B</fnm></au>
    <au><snm>Yu</snm><fnm>PS</fnm></au>
    <au><snm>Zhou</snm><fnm>ZH</fnm></au>
    <au><snm>Steinbach</snm><fnm>M</fnm></au>
    <au><snm>Hand</snm><fnm>DJ</fnm></au>
    <au><snm>Steinberg</snm><fnm>D</fnm></au>
  </aug>
  <source>Knowledge and Information Systems</source>
  <publisher>Berlin, Heidelberg: Springer-Verlag</publisher>
  <pubdate>2007</pubdate>
  <volume>14</volume>
  <issue>1</issue>
  <fpage>1–37</fpage>
</bibl>

<bibl id="B52">
  <title><p>Intel SGX Explained</p></title>
  <aug>
    <au><snm>Costan</snm><fnm>V</fnm></au>
    <au><snm>Devadas</snm><fnm>S</fnm></au>
  </aug>
  <source>IACR Cryptology ePrint Archive</source>
  <pubdate>2016</pubdate>
  <volume>2016</volume>
  <fpage>86</fpage>
</bibl>

<bibl id="B53">
  <title><p>A Brief Introduction to Boosting</p></title>
  <aug>
    <au><snm>Schapire</snm><fnm>RE</fnm></au>
  </aug>
  <source>Proceedings of the 16th International Joint Conference on Artificial
  Intelligence - Volume 2</source>
  <publisher>San Francisco, CA, USA: Morgan Kaufmann Publishers
  Inc.</publisher>
  <series><title><p>IJCAI'99</p></title></series>
  <pubdate>1999</pubdate>
  <fpage>1401–1406</fpage>
</bibl>

<bibl id="B54">
  <title><p>Deep learning</p></title>
  <aug>
    <au><snm>LeCun</snm><fnm>Y</fnm></au>
    <au><snm>Bengio</snm><fnm>Y</fnm></au>
    <au><snm>Hinton</snm><fnm>G</fnm></au>
  </aug>
  <source>Nature</source>
  <pubdate>2015</pubdate>
  <volume>521</volume>
  <fpage>436–</fpage>
  <lpage>444</lpage>
</bibl>

<bibl id="B55">
  <title><p>Privacy-Preserving Deep Learning via Additively Homomorphic
  Encryption</p></title>
  <aug>
    <au><snm>{Phong}</snm><fnm>L. T.</fnm></au>
    <au><snm>{Aono}</snm><fnm>Y.</fnm></au>
    <au><snm>{Hayashi}</snm><fnm>T.</fnm></au>
    <au><snm>{Wang}</snm><fnm>L.</fnm></au>
    <au><snm>{Moriai}</snm><fnm>S.</fnm></au>
  </aug>
  <source>IEEE Transactions on Information Forensics and Security</source>
  <pubdate>2018</pubdate>
  <volume>13</volume>
  <issue>5</issue>
  <fpage>1333</fpage>
  <lpage>1345</lpage>
</bibl>

<bibl id="B56">
  <title><p>Deepsecure: Scalable Provably-Secure Deep Learning</p></title>
  <aug>
    <au><snm>Rouhani</snm><fnm>BD</fnm></au>
    <au><snm>Riazi</snm><fnm>MS</fnm></au>
    <au><snm>Koushanfar</snm><fnm>F</fnm></au>
  </aug>
  <source>Proceedings of the 55th Annual Design Automation Conference</source>
  <publisher>New York, NY, USA: Association for Computing Machinery</publisher>
  <series><title><p>DAC '18</p></title></series>
  <pubdate>2018</pubdate>
  <url>https://doi.org/10.1145/3195970.3196023</url>
</bibl>

<bibl id="B57">
  <title><p>Evaluation of the Complexity of Fully Homomorphic Encryption
  Schemes in Implementations of Programs</p></title>
  <aug>
    <au><snm>Chakarov</snm><fnm>D</fnm></au>
    <au><snm>Papazov</snm><fnm>Y</fnm></au>
  </aug>
  <source>Proceedings of the 20th International Conference on Computer Systems
  and Technologies</source>
  <publisher>New York, NY, USA: Association for Computing Machinery</publisher>
  <series><title><p>CompSysTech '19</p></title></series>
  <pubdate>2019</pubdate>
  <fpage>62?67</fpage>
  <url>https://doi.org/10.1145/3345252.3345292</url>
</bibl>

<bibl id="B58">
  <title><p>Encrypted integer division and secure comparison</p></title>
  <aug>
    <au><snm>Veugen</snm><fnm>T</fnm></au>
  </aug>
  <source>International Journal of Applied Cryptography</source>
  <pubdate>2014</pubdate>
  <volume>3</volume>
  <issue>2</issue>
  <fpage>166</fpage>
</bibl>

<bibl id="B59">
  <title><p>A private stable matching algorithm</p></title>
  <aug>
    <au><snm>Golle</snm><fnm>P</fnm></au>
  </aug>
  <source>International Conference on Financial Cryptography and Data
  Security</source>
  <pubdate>2006</pubdate>
  <fpage>65</fpage>
  <lpage>-80</lpage>
</bibl>

<bibl id="B60">
  <title><p>How to Play ANY Mental Game</p></title>
  <aug>
    <au><snm>Goldreich</snm><fnm>O.</fnm></au>
    <au><snm>Micali</snm><fnm>S.</fnm></au>
    <au><snm>Wigderson</snm><fnm>A.</fnm></au>
  </aug>
  <source>Proceedings of the Nineteenth Annual ACM Symposium on Theory of
  Computing</source>
  <publisher>New York, NY, USA: ACM</publisher>
  <series><title><p>STOC '87</p></title></series>
  <pubdate>1987</pubdate>
  <fpage>218</fpage>
  <lpage>-229</lpage>
  <url>http://doi.acm.org/10.1145/28395.28420</url>
</bibl>

<bibl id="B61">
  <title><p>Secure Two-party K-means Clustering</p></title>
  <aug>
    <au><snm>Bunn</snm><fnm>P</fnm></au>
    <au><snm>Ostrovsky</snm><fnm>R</fnm></au>
  </aug>
  <source>Proceedings of the 14th ACM Conference on Computer and Communications
  Security</source>
  <publisher>New York, NY, USA: ACM</publisher>
  <series><title><p>CCS '07</p></title></series>
  <pubdate>2007</pubdate>
  <fpage>486</fpage>
  <lpage>-497</lpage>
  <url>http://doi.acm.org/10.1145/1315245.1315306</url>
</bibl>

<bibl id="B62">
  <title><p>Privacy preserving string comparisons based on Levenshtein
  distance</p></title>
  <aug>
    <au><snm>Rane</snm><fnm>S.</fnm></au>
    <au><snm>Sun</snm><fnm>W.</fnm></au>
  </aug>
  <source>2010 IEEE International Workshop on Information Forensics and
  Security</source>
  <pubdate>2010</pubdate>
  <fpage>1</fpage>
  <lpage>6</lpage>
</bibl>

<bibl id="B63">
  <title><p>Division between encrypted integers by means of Garbled
  Circuits</p></title>
  <aug>
    <au><snm>Lazzeretti</snm><fnm>R.</fnm></au>
    <au><snm>Barni</snm><fnm>M.</fnm></au>
  </aug>
  <source>2011 IEEE International Workshop on Information Forensics and
  Security</source>
  <pubdate>2011</pubdate>
  <fpage>1</fpage>
  <lpage>6</lpage>
</bibl>

<bibl id="B64">
  <title><p>On Secure Two-Party Integer Division</p></title>
  <aug>
    <au><snm>Dahl</snm><fnm>M</fnm></au>
    <au><snm>Ning</snm><fnm>C</fnm></au>
    <au><snm>Toft</snm><fnm>T</fnm></au>
  </aug>
  <source>Financial Cryptography and Data Security</source>
  <publisher>Berlin, Heidelberg: Springer Berlin Heidelberg</publisher>
  <editor>Keromytis, Angelos D.</editor>
  <pubdate>2012</pubdate>
  <fpage>164</fpage>
  <lpage>-178</lpage>
</bibl>

<bibl id="B65">
  <title><p>Secure Outsourced Matrix Computation and Application to Neural
  Networks</p></title>
  <aug>
    <au><snm>Jiang</snm><fnm>X</fnm></au>
    <au><snm>Kim</snm><fnm>M</fnm></au>
    <au><snm>Lauter</snm><fnm>K</fnm></au>
    <au><snm>Song</snm><fnm>Y</fnm></au>
  </aug>
  <source>Proceedings of the 2018 ACM SIGSAC Conference on Computer and
  Communications Security</source>
  <publisher>New York, NY, USA: ACM</publisher>
  <series><title><p>CCS '18</p></title></series>
  <pubdate>2018</pubdate>
  <fpage>1209</fpage>
  <lpage>-1222</lpage>
  <url>http://doi.acm.org/10.1145/3243734.3243837</url>
</bibl>

<bibl id="B66">
  <title><p>Design and implementation of a homomorphic-encryption
  library</p></title>
  <aug>
    <au><snm>Halevi</snm><fnm>S</fnm></au>
    <au><snm>Shoup</snm><fnm>V</fnm></au>
  </aug>
  <pubdate>2013</pubdate>
</bibl>

<bibl id="B67">
  <title><p>Chameleon: A Hybrid Secure Computation Framework for Machine
  Learning Applications</p></title>
  <aug>
    <au><snm>Riazi</snm><fnm>MS</fnm></au>
    <au><snm>Weinert</snm><fnm>C</fnm></au>
    <au><snm>Tkachenko</snm><fnm>O</fnm></au>
    <au><snm>Songhori</snm><fnm>EM</fnm></au>
    <au><snm>Schneider</snm><fnm>T</fnm></au>
    <au><snm>Koushanfar</snm><fnm>F</fnm></au>
  </aug>
  <source>Proceedings of the 2018 on Asia Conference on Computer and
  Communications Security</source>
  <publisher>New York, NY, USA: Association for Computing Machinery</publisher>
  <series><title><p>ASIACCS '18</p></title></series>
  <pubdate>2018</pubdate>
  <fpage>707?721</fpage>
  <url>https://doi.org/10.1145/3196494.3196522</url>
</bibl>

<bibl id="B68">
  <title><p>ABY3: A Mixed Protocol Framework for Machine Learning</p></title>
  <aug>
    <au><snm>Mohassel</snm><fnm>P</fnm></au>
    <au><snm>Rindal</snm><fnm>P</fnm></au>
  </aug>
  <source>Cryptology ePrint Archive, Report 2018/403</source>
  <pubdate>2018</pubdate>
  <note>\url{https://eprint.iacr.org/2018/403}</note>
</bibl>

<bibl id="B69">
  <title><p>BLAZE: Blazing Fast Privacy-Preserving Machine Learning</p></title>
  <aug>
    <au><snm>Patra</snm><fnm>A</fnm></au>
    <au><snm>Suresh</snm><fnm>A</fnm></au>
  </aug>
  <source>Cryptology ePrint Archive, Report 2020/042</source>
  <pubdate>2020</pubdate>
  <note>\url{https://eprint.iacr.org/2020/042}</note>
</bibl>

<bibl id="B70">
  <title><p>Practical Privacy-Preserving Multiparty Linear Programming Based on
  Problem Transformation</p></title>
  <aug>
    <au><snm>Dreier</snm><fnm>J</fnm></au>
    <au><snm>Kerschbaum</snm><fnm>F</fnm></au>
  </aug>
  <source>Proceedings of the third IEEE Conference on Social Computing</source>
  <pubdate>2011</pubdate>
  <fpage>916</fpage>
  <lpage>924</lpage>
</bibl>

<bibl id="B71">
  <title><p>TASTY: Tool for Automating Secure Two-party
  Computations</p></title>
  <aug>
    <au><snm>Henecka</snm><fnm>W</fnm></au>
    <au><snm>K \"{o}gl</snm><fnm>S</fnm></au>
    <au><snm>Sadeghi</snm><fnm>AR</fnm></au>
    <au><snm>Schneider</snm><fnm>T</fnm></au>
    <au><snm>Wehrenberg</snm><fnm>I</fnm></au>
  </aug>
  <source>Proceedings of the 17th ACM Conference on Computer and Communications
  Security</source>
  <publisher>New York, NY, USA: ACM</publisher>
  <series><title><p>CCS '10</p></title></series>
  <pubdate>2010</pubdate>
  <fpage>451</fpage>
  <lpage>-462</lpage>
  <url>http://doi.acm.org/10.1145/1866307.1866358</url>
</bibl>

<bibl id="B72">
  <title><p>Can homomorphic encryption be practical?</p></title>
  <aug>
    <au><snm>Naehrig</snm><fnm>M</fnm></au>
    <au><snm>Lauter</snm><fnm>K</fnm></au>
    <au><snm>Vaikuntanathan</snm><fnm>V</fnm></au>
  </aug>
  <source>Proceedings of cloud computing security workshop</source>
  <publisher>New York, NY, USA: ACM</publisher>
  <pubdate>2011</pubdate>
  <fpage>113</fpage>
  <lpage>-124</lpage>
</bibl>

<bibl id="B73">
  <title><p>Universally Composable Security</p></title>
  <aug>
    <au><snm>Canetti</snm><fnm>R</fnm></au>
  </aug>
  <source>J. ACM</source>
  <publisher>New York, NY, USA: Association for Computing Machinery</publisher>
  <pubdate>2020</pubdate>
  <volume>67</volume>
  <issue>5</issue>
</bibl>

<bibl id="B74">
  <title><p>How to Simulate It -- A Tutorial on the Simulation Proof
  Technique</p></title>
  <aug>
    <au><snm>Lindell</snm><fnm>Y</fnm></au>
  </aug>
  <source>Tutorials on the Foundations of Cryptography: Dedicated to Oded
  Goldreich</source>
  <publisher>Cham: Springer International Publishing</publisher>
  <editor>Lindell, Yehuda</editor>
  <pubdate>2017</pubdate>
  <fpage>277</fpage>
  <lpage>-346</lpage>
</bibl>

<bibl id="B75">
  <title><p>Stealing Machine Learning Models via Prediction APIs</p></title>
  <aug>
    <au><snm>Tram\`{e}r</snm><fnm>F</fnm></au>
    <au><snm>Zhang</snm><fnm>F</fnm></au>
    <au><snm>Juels</snm><fnm>A</fnm></au>
    <au><snm>Reiter</snm><fnm>MK</fnm></au>
    <au><snm>Ristenpart</snm><fnm>T</fnm></au>
  </aug>
  <source>Proceedings of the 25th USENIX Conference on Security
  Symposium</source>
  <publisher>USA: USENIX Association</publisher>
  <series><title><p>SEC'16</p></title></series>
  <pubdate>2016</pubdate>
  <fpage>601–618</fpage>
</bibl>

<bibl id="B76">
  <title><p>MapReduce: Simplified Data Processing on Large Clusters</p></title>
  <aug>
    <au><snm>Dean</snm><fnm>J</fnm></au>
    <au><snm>Ghemawat</snm><fnm>S</fnm></au>
  </aug>
  <source>OSDI</source>
  <pubdate>2004</pubdate>
  <fpage>137</fpage>
  <lpage>150</lpage>
</bibl>

<bibl id="B77">
  <title><p>RAPPOR: Randomized Aggregatable Privacy-Preserving Ordinal
  Response</p></title>
  <aug>
    <au><snm>Erlingsson</snm><fnm>U</fnm></au>
    <au><snm>Pihur</snm><fnm>V</fnm></au>
    <au><snm>Korolova</snm><fnm>A</fnm></au>
  </aug>
  <source>Proceedings of the 2014 ACM SIGSAC Conference on Computer and
  Communications Security</source>
  <publisher>New York, NY, USA: ACM</publisher>
  <series><title><p>CCS '14</p></title></series>
  <pubdate>2014</pubdate>
  <fpage>1054</fpage>
  <lpage>-1067</lpage>
  <url>http://doi.acm.org/10.1145/2660267.2660348</url>
</bibl>

<bibl id="B78">
  <title><p>Using Randomized Response Techniques for Privacy-preserving Data
  Mining</p></title>
  <aug>
    <au><snm>Du</snm><fnm>W</fnm></au>
    <au><snm>Zhan</snm><fnm>Z</fnm></au>
  </aug>
  <source>Proceedings of the Ninth ACM SIGKDD International Conference on
  Knowledge Discovery and Data Mining</source>
  <publisher>ACM</publisher>
  <pubdate>2003</pubdate>
  <fpage>505</fpage>
  <lpage>-510</lpage>
</bibl>

<bibl id="B79">
  <title><p>Privacy-Preserving Data Mining</p></title>
  <aug>
    <au><snm>Agrawal</snm><fnm>R</fnm></au>
    <au><snm>Srikant</snm><fnm>R</fnm></au>
  </aug>
  <source>{Proceedings of ACM SIGMOD Conference}</source>
  <publisher>Dallas, Texas: ACM</publisher>
  <pubdate>2000</pubdate>
  <fpage>439</fpage>
  <lpage>-450</lpage>
</bibl>

<bibl id="B80">
  <title><p>Geometric Data Perturbation for Outsourced Data Mining</p></title>
  <aug>
    <au><snm>Chen</snm><fnm>K</fnm></au>
    <au><snm>Liu</snm><fnm>L</fnm></au>
  </aug>
  <source>{Knowledge and Information Systems}</source>
  <pubdate>2011</pubdate>
  <volume>29</volume>
  <issue>3</issue>
</bibl>

<bibl id="B81">
  <title><p>Random Projection-Based Multiplicative Data Perturbation for
  Privacy Preserving Distributed Data Mining</p></title>
  <aug>
    <au><snm>Liu</snm><fnm>K</fnm></au>
    <au><snm>Kargupta</snm><fnm>H</fnm></au>
    <au><snm>Ryan</snm><fnm>J</fnm></au>
  </aug>
  <source>IEEE Transactions on Knowledge and Data Engineering (TKDE)</source>
  <pubdate>2006</pubdate>
  <volume>18</volume>
  <issue>1</issue>
  <fpage>92</fpage>
  <lpage>106</lpage>
</bibl>

<bibl id="B82">
  <title><p>Building Confidential and Efficient Query Services in the Cloud
  with RASP Data Perturbation</p></title>
  <aug>
    <au><snm>Xu</snm><fnm>H</fnm></au>
    <au><snm>Guo</snm><fnm>S</fnm></au>
    <au><snm>Chen</snm><fnm>K</fnm></au>
  </aug>
  <source>{IEEE Transactions on Knowledge and Data Engineering}</source>
  <pubdate>2014</pubdate>
  <volume>26</volume>
  <issue>2</issue>
</bibl>

<bibl id="B83">
  <title><p>SgxPectre Attacks: Leaking Enclave Secrets via Speculative
  Execution</p></title>
  <aug>
    <au><snm>Chen</snm><fnm>G</fnm></au>
    <au><snm>Chen</snm><fnm>S</fnm></au>
    <au><snm>Xiao</snm><fnm>Y</fnm></au>
    <au><snm>Zhang</snm><fnm>Y</fnm></au>
    <au><snm>Lin</snm><fnm>Z</fnm></au>
    <au><snm>Lai</snm><fnm>TH</fnm></au>
  </aug>
  <source>CoRR</source>
  <pubdate>2018</pubdate>
  <volume>abs/1802.09085</volume>
</bibl>

<bibl id="B84">
  <title><p>Image Disguising for Privacy-Preserving Outsourced Deep
  Learning</p></title>
  <aug>
    <au><snm>Sharma</snm><fnm>S</fnm></au>
    <au><snm>Chen</snm><fnm>K</fnm></au>
  </aug>
  <source>Poster Session of ACM CCS</source>
  <pubdate>2018</pubdate>
</bibl>

<bibl id="B85">
  <title><p>Order-Preserving Encryption Revisited:Improved Security Analysisand
  Alternative Solutions</p></title>
  <aug>
    <au><snm>Boldyreva</snm><fnm>A</fnm></au>
    <au><snm>Chenette</snm><fnm>N</fnm></au>
    <au><snm>O'Neill</snm><fnm>A</fnm></au>
  </aug>
  <source>CRYPTO</source>
  <pubdate>2011</pubdate>
</bibl>

<bibl id="B86">
  <title><p>Order Preserving Symmetric Encryption</p></title>
  <aug>
    <au><snm>Boldyreva</snm><fnm>A</fnm></au>
    <au><snm>Chenette</snm><fnm>N</fnm></au>
    <au><snm>Lee</snm><fnm>Y</fnm></au>
    <au><snm>O'Neill</snm><fnm>A</fnm></au>
  </aug>
  <source>Proceedings of EUROCRYPT conference</source>
  <pubdate>2009</pubdate>
</bibl>

<bibl id="B87">
  <title><p>Frequency-Hiding Order-Preserving Encryption</p></title>
  <aug>
    <au><snm>Kerschbaum</snm><fnm>F</fnm></au>
  </aug>
  <source>Proceedings of ACM Conference on Computer and Communication
  Security</source>
  <pubdate>2015</pubdate>
</bibl>

<bibl id="B88">
  <title><p>Secure Conjunctive Keyword Search over Encrypted Data</p></title>
  <aug>
    <au><snm>Golle</snm><fnm>P</fnm></au>
    <au><snm>Staddon</snm><fnm>J</fnm></au>
    <au><snm>Waters</snm><fnm>B</fnm></au>
  </aug>
  <source>ACNS 04: 2nd International Conference on Applied Cryptography and
  Network Security</source>
  <publisher>Springer-Verlag</publisher>
  <pubdate>2004</pubdate>
  <fpage>31</fpage>
  <lpage>-45</lpage>
</bibl>

<bibl id="B89">
  <title><p>Searchable symmetric encryption: improved definitions and efficient
  constructions</p></title>
  <aug>
    <au><snm>Curtmola</snm><fnm>R</fnm></au>
    <au><snm>Garay</snm><fnm>J</fnm></au>
    <au><snm>Kamara</snm><fnm>S</fnm></au>
    <au><snm>Ostrovsky</snm><fnm>R</fnm></au>
  </aug>
  <source>ACM CCS</source>
  <pubdate>2006</pubdate>
  <fpage>79</fpage>
  <lpage>-88</lpage>
</bibl>

<bibl id="B90">
  <title><p>Graphene-SGX: {A} Practical Library {OS} for Unmodified
  Applications on {SGX}</p></title>
  <aug>
    <au><snm>Tsai</snm><fnm>C</fnm></au>
    <au><snm>Porter</snm><fnm>DE</fnm></au>
    <au><snm>Vij</snm><fnm>M</fnm></au>
  </aug>
  <source>2017 {USENIX} Annual Technical Conference, {USENIX} {ATC} 2017, Santa
  Clara, CA, USA, July 12-14, 2017</source>
  <editor>Dilma Da Silva and Bryan Ford</editor>
  <pubdate>2017</pubdate>
  <fpage>645</fpage>
  <lpage>-658</lpage>
</bibl>

<bibl id="B91">
  <title><p>SCONE: Secure Linux Containers with Intel SGX</p></title>
  <aug>
    <au><snm>Arnautov</snm><fnm>S</fnm></au>
    <au><snm>Trach</snm><fnm>B</fnm></au>
    <au><snm>Gregor</snm><fnm>F</fnm></au>
    <au><snm>Knauth</snm><fnm>T</fnm></au>
    <au><snm>Martin</snm><fnm>A</fnm></au>
    <au><snm>Priebe</snm><fnm>C</fnm></au>
    <au><snm>Lind</snm><fnm>J</fnm></au>
    <au><snm>Muthukumaran</snm><fnm>D</fnm></au>
    <au><snm>O'Keeffe</snm><fnm>D</fnm></au>
    <au><snm>Stillwell</snm><fnm>ML</fnm></au>
    <au><snm>Goltzsche</snm><fnm>D</fnm></au>
    <au><snm>Eyers</snm><fnm>D</fnm></au>
    <au><snm>Kapitza</snm><fnm>R</fnm></au>
    <au><snm>Pietzuch</snm><fnm>P</fnm></au>
    <au><snm>Fetzer</snm><fnm>C</fnm></au>
  </aug>
  <source>Proceedings of the 12th USENIX Conference on Operating Systems Design
  and Implementation</source>
  <publisher>Berkeley, CA, USA: USENIX Association</publisher>
  <series><title><p>OSDI'16</p></title></series>
  <pubdate>2016</pubdate>
  <fpage>689</fpage>
  <lpage>-703</lpage>
</bibl>

<bibl id="B92">
  <title><p>Panoply: Low-TCB Linux Applications With SGX Enclaves</p></title>
  <aug>
    <au><snm>Shinde</snm><fnm>S</fnm></au>
    <au><snm>Tien</snm><fnm>DL</fnm></au>
    <au><snm>Tople</snm><fnm>S</fnm></au>
    <au><snm>Saxena</snm><fnm>P</fnm></au>
  </aug>
  <source>Proceedings of NDSS</source>
  <pubdate>2017</pubdate>
</bibl>

<bibl id="B93">
  <title><p>Privacy-Preserving Machine Learning in Untrusted Clouds Made
  Simple</p></title>
  <aug>
    <au><snm>Lee</snm><fnm>D</fnm></au>
    <au><snm>Kuvaiskii</snm><fnm>D</fnm></au>
    <au><snm>Vahldiek{-}Oberwagner</snm><fnm>A</fnm></au>
    <au><snm>Vij</snm><fnm>M</fnm></au>
  </aug>
  <source>CoRR</source>
  <pubdate>2020</pubdate>
  <volume>abs/2009.04390</volume>
  <url>https://arxiv.org/abs/2009.04390</url>
</bibl>

<bibl id="B94">
  <title><p>ZeroTrace : Oblivious Memory Primitives from Intel
  {SGX}</p></title>
  <aug>
    <au><snm>Sasy</snm><fnm>S</fnm></au>
    <au><snm>Gorbunov</snm><fnm>S</fnm></au>
    <au><snm>Fletcher</snm><fnm>CW</fnm></au>
  </aug>
  <source>25th Annual Network and Distributed System Security Symposium, {NDSS}
  2018, San Diego, California, USA, February 18-21, 2018</source>
  <pubdate>2018</pubdate>
</bibl>

<bibl id="B95">
  <title><p>OBLIVIATE: A Data Oblivious File System for Intel SGX</p></title>
  <aug>
    <au><snm>Ahmad</snm><fnm>A</fnm></au>
    <au><snm>Kim</snm><fnm>K</fnm></au>
    <au><snm>Sarfaraz</snm><fnm>MI</fnm></au>
    <au><snm>Lee</snm><fnm>B</fnm></au>
  </aug>
  <source>the Network and Distributed System Security Symposium</source>
  <pubdate>2018</pubdate>
</bibl>

<bibl id="B96">
  <title><p>Preventing Page Faults from Telling Your Secrets</p></title>
  <aug>
    <au><snm>Shinde</snm><fnm>S</fnm></au>
    <au><snm>Chua</snm><fnm>ZL</fnm></au>
    <au><snm>Narayanan</snm><fnm>V</fnm></au>
    <au><snm>Saxena</snm><fnm>P</fnm></au>
  </aug>
  <source>Proceedings of the 11th ACM on Asia Conference on Computer and
  Communications Security</source>
  <publisher>New York, NY, USA: Association for Computing Machinery</publisher>
  <series><title><p>ASIACCS16</p></title></series>
  <pubdate>2016</pubdate>
  <fpage>317–328</fpage>
  <url>https://doi.org/10.1145/2897845.2897885</url>
</bibl>

<bibl id="B97">
  <title><p>Software Protection and Simulation on Oblivious RAM</p></title>
  <aug>
    <au><snm>Goldreich</snm><fnm>O</fnm></au>
    <au><snm>Ostrovsky</snm><fnm>R</fnm></au>
  </aug>
  <source>Journal of the ACM</source>
  <pubdate>1996</pubdate>
  <volume>43</volume>
  <fpage>431</fpage>
  <lpage>-473</lpage>
</bibl>

<bibl id="B98">
  <title><p>Oblivious Multi-Party Machine Learning on Trusted
  Processors</p></title>
  <aug>
    <au><snm>Ohrimenko</snm><fnm>O</fnm></au>
    <au><snm>Schuster</snm><fnm>F</fnm></au>
    <au><snm>Fournet</snm><fnm>C</fnm></au>
    <au><snm>Mehta</snm><fnm>A</fnm></au>
    <au><snm>Nowozin</snm><fnm>S</fnm></au>
    <au><snm>Vaswani</snm><fnm>K</fnm></au>
    <au><snm>Costa</snm><fnm>M</fnm></au>
  </aug>
  <source>25th {USENIX} Security Symposium, {USENIX} Security 16, Austin, TX,
  USA, August 10-12, 2016</source>
  <publisher>{USENIX} Association</publisher>
  <editor>Thorsten Holz and Stefan Savage</editor>
  <pubdate>2016</pubdate>
  <fpage>619</fpage>
  <lpage>-636</lpage>
  <url>https://www.usenix.org/conference/usenixsecurity16/technical-sessions/presentation/ohrimenko</url>
</bibl>

<bibl id="B99">
  <title><p>SGX-MR: Regulating Dataflows for Protecting Access Patterns of
  Data-Intensive SGX Applications</p></title>
  <aug>
    <au><snm>Alam</snm><fnm>AKMM</fnm></au>
    <au><snm>Sharma</snm><fnm>S</fnm></au>
    <au><snm>Chen</snm><fnm>K</fnm></au>
  </aug>
  <source>Proceedings on Privacy Enhancing Technologies</source>
  <publisher>Berlin: Sciendo</publisher>
  <pubdate>01 Jan. 2021</pubdate>
  <volume>2021</volume>
  <issue>1</issue>
  <fpage>5</fpage>
  <lpage>20</lpage>
  <url>https://content.sciendo.com/view/journals/popets/2021/1/article-p5.xml</url>
</bibl>

<bibl id="B100">
  <title><p>Foreshadow: Extracting the Keys to the Intel {SGX} Kingdom with
  Transient Out-of-Order Execution</p></title>
  <aug>
    <au><snm>Bulck</snm><fnm>JV</fnm></au>
    <au><snm>Minkin</snm><fnm>M</fnm></au>
    <au><snm>Weisse</snm><fnm>O</fnm></au>
    <au><snm>Genkin</snm><fnm>D</fnm></au>
    <au><snm>Kasikci</snm><fnm>B</fnm></au>
    <au><snm>Piessens</snm><fnm>F</fnm></au>
    <au><snm>Silberstein</snm><fnm>M</fnm></au>
    <au><snm>Wenisch</snm><fnm>TF</fnm></au>
    <au><snm>Yarom</snm><fnm>Y</fnm></au>
    <au><snm>Strackx</snm><fnm>R</fnm></au>
  </aug>
  <source>27th {USENIX} Security Symposium ({USENIX} Security 18)</source>
  <publisher>Baltimore, MD: {USENIX} Association</publisher>
  <pubdate>2018</pubdate>
  <fpage>991{\textendash}1008</fpage>
  <url>https://www.usenix.org/conference/usenixsecurity18/presentation/bulck</url>
</bibl>

<bibl id="B101">
  <title><p>Hey, you, get off of my cloud: exploring information leakage in
  third-party compute clouds</p></title>
  <aug>
    <au><snm>Ristenpart</snm><fnm>T</fnm></au>
    <au><snm>Tromer</snm><fnm>E</fnm></au>
    <au><snm>Shacham</snm><fnm>H</fnm></au>
    <au><snm>Savage</snm><fnm>S</fnm></au>
  </aug>
  <source>Proceedings of the 16th ACM conference on Computer and communications
  security</source>
  <publisher>New York, NY, USA</publisher>
  <pubdate>2009</pubdate>
  <fpage>199</fpage>
  <lpage>-212</lpage>
</bibl>

<bibl id="B102">
  <title><p>Spectre Attacks: Exploiting Speculative Execution</p></title>
  <aug>
    <au><snm>{Kocher}</snm><fnm>P.</fnm></au>
    <au><snm>{Horn}</snm><fnm>J.</fnm></au>
    <au><snm>{Fogh}</snm><fnm>A.</fnm></au>
    <au><snm>{Genkin}</snm><fnm>D.</fnm></au>
    <au><snm>{Gruss}</snm><fnm>D.</fnm></au>
    <au><snm>{Haas}</snm><fnm>W.</fnm></au>
    <au><snm>{Hamburg}</snm><fnm>M.</fnm></au>
    <au><snm>{Lipp}</snm><fnm>M.</fnm></au>
    <au><snm>{Mangard}</snm><fnm>S.</fnm></au>
    <au><snm>{Prescher}</snm><fnm>T.</fnm></au>
    <au><snm>{Schwarz}</snm><fnm>M.</fnm></au>
    <au><snm>{Yarom}</snm><fnm>Y.</fnm></au>
  </aug>
  <source>2019 IEEE Symposium on Security and Privacy (SP)</source>
  <pubdate>2019</pubdate>
  <fpage>1</fpage>
  <lpage>19</lpage>
</bibl>

<bibl id="B103">
  <title><p>Meltdown: Reading Kernel Memory from User Space</p></title>
  <aug>
    <au><snm>Lipp</snm><fnm>M</fnm></au>
    <au><snm>Schwarz</snm><fnm>M</fnm></au>
    <au><snm>Gruss</snm><fnm>D</fnm></au>
    <au><snm>Prescher</snm><fnm>T</fnm></au>
    <au><snm>Haas</snm><fnm>W</fnm></au>
    <au><snm>Fogh</snm><fnm>A</fnm></au>
    <au><snm>Horn</snm><fnm>J</fnm></au>
    <au><snm>Mangard</snm><fnm>S</fnm></au>
    <au><snm>Kocher</snm><fnm>P</fnm></au>
    <au><snm>Genkin</snm><fnm>D</fnm></au>
    <au><snm>Yarom</snm><fnm>Y</fnm></au>
    <au><snm>Hamburg</snm><fnm>M</fnm></au>
  </aug>
  <source>27th {USENIX} Security Symposium ({USENIX} Security 18)</source>
  <pubdate>2018</pubdate>
</bibl>

</refgrp>
} 


\newpage 
\section*{Figures}

\begin{figure}[h!]  
 \centering 
 \includegraphics[width=0.7\linewidth]{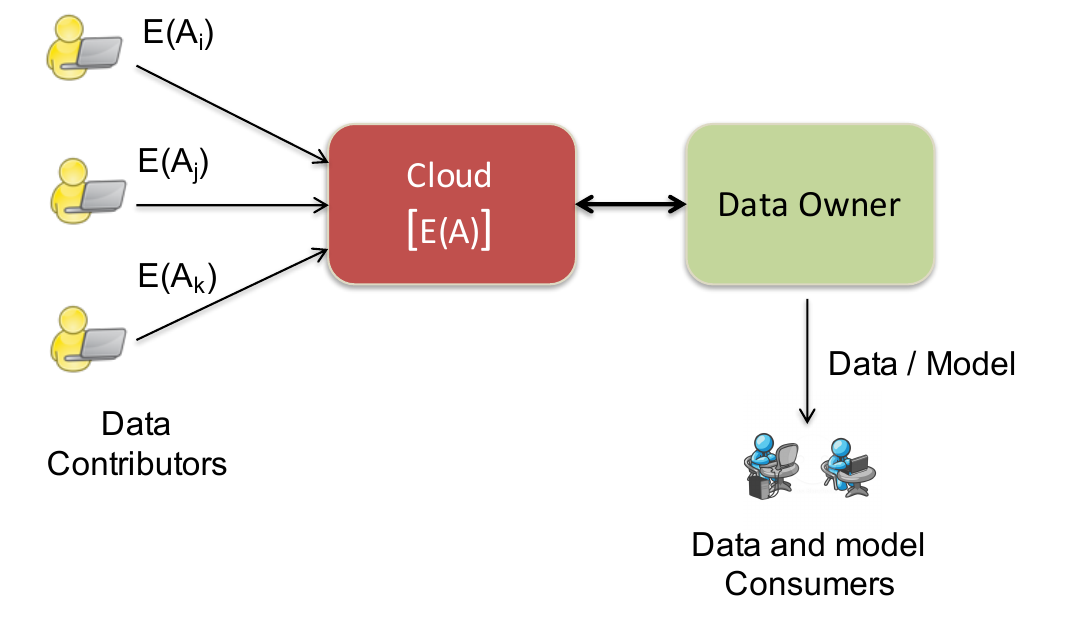}
\caption{A  data owner outsourcing to an untrusted cloud provider for learning a model. The data contributors directly submit their encrypted data to the cloud.  The cloud carries out the major expensive computations over the encrypted data and data owner can assist with some lightweight work.}
\label{fig:hbc_CP}
\end{figure}

\begin{figure}[h!] 
 \centering 
 \includegraphics[width=0.8\linewidth]{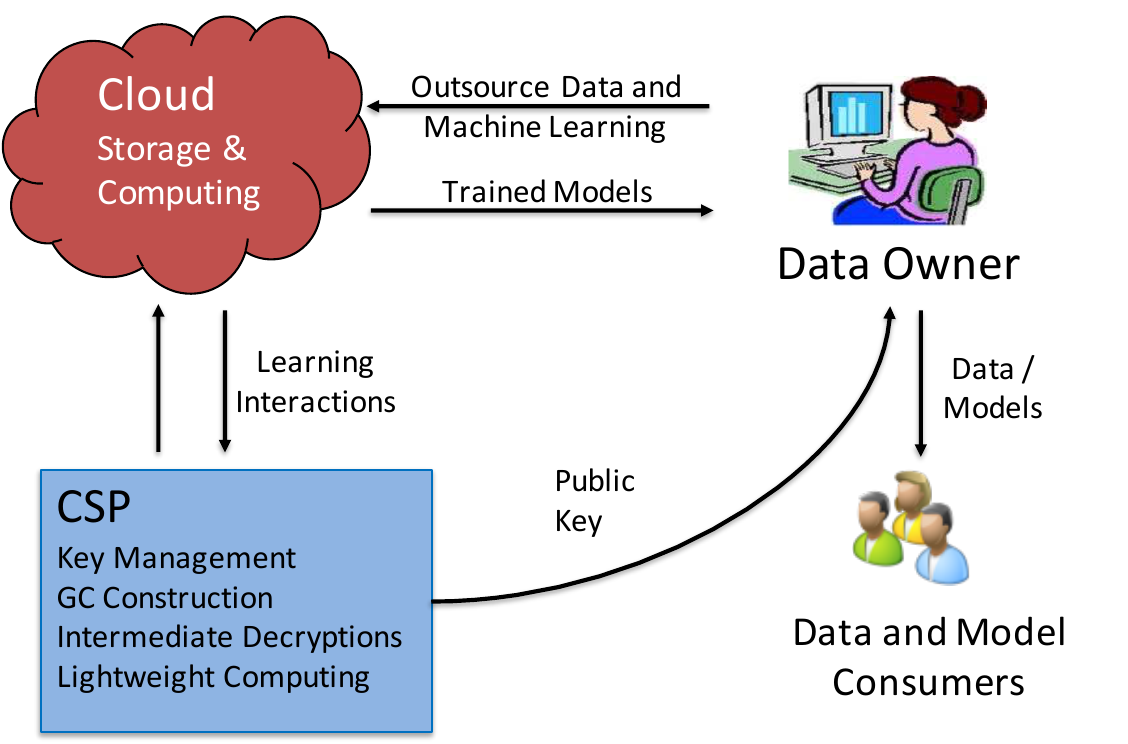}
\caption{A  data owner outsources data storage and machine learning tasks to the Cloud. The Cryptographic Service Provider (CSP) manages the keys, decrypts intermediate results, and assists the Cloud with other relatively lightweight computations.}
\label{fig:parties}
\end{figure}

\begin{sidewaysfigure}[htb]  
 \centering 
 \includegraphics[width=1.0\linewidth]{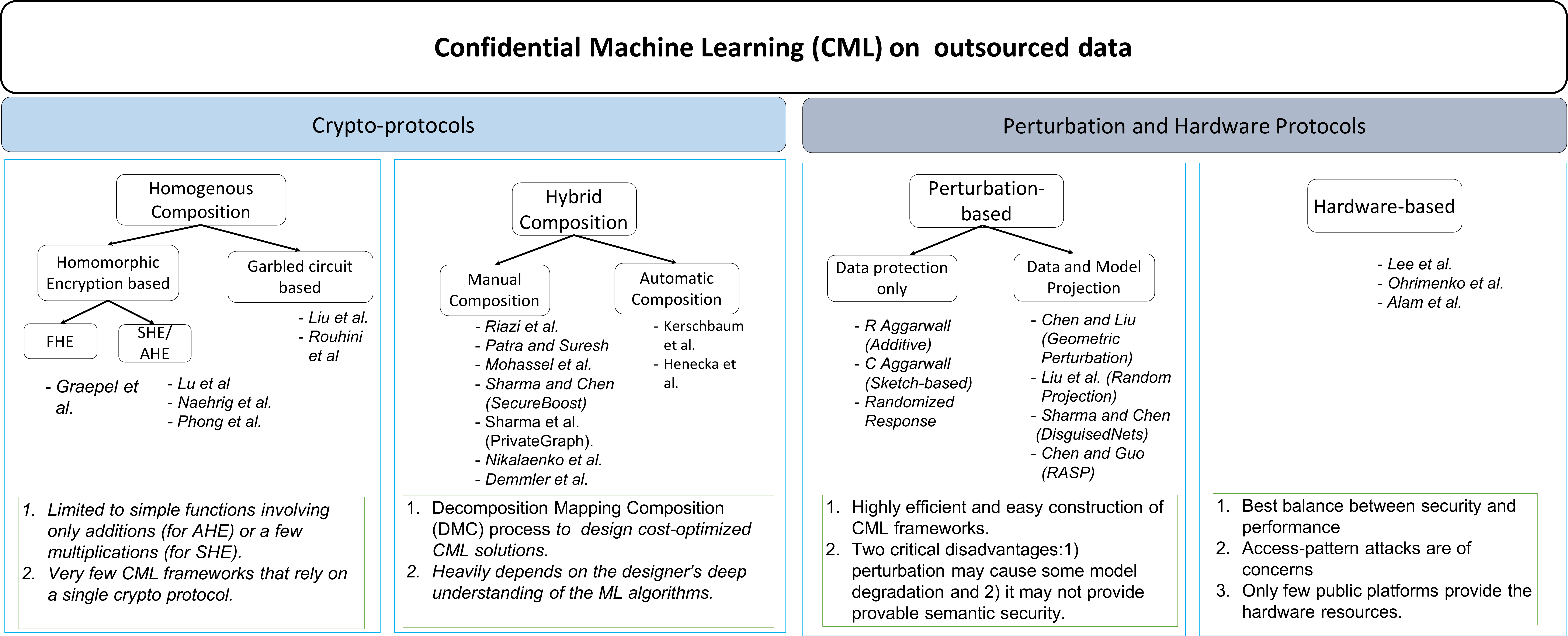}
\caption{The systematization framework for confidential machine learning (CML) approaches.}
\label{fig:system}
\end{sidewaysfigure}

\begin{figure}[h!]  
 \centering 
 \includegraphics[width=1.0\linewidth]{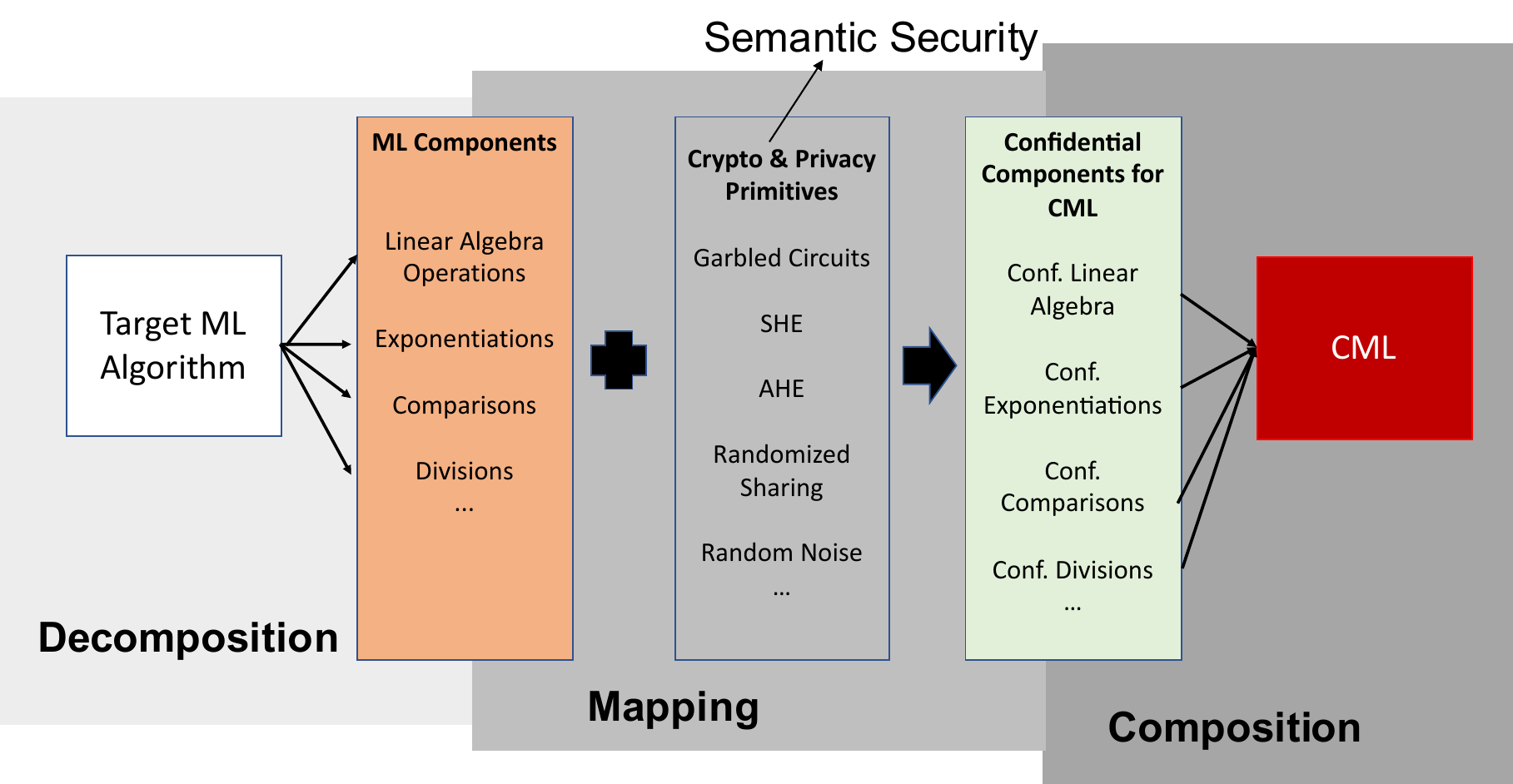}
\caption{The decomposition-mapping-composition (DMC) process for constructing hybrid CML solutions.}
\label{fig:dmc}
\end{figure}


\newpage

\section*{Tables}
\begin{table} [h!] 
\centering
  \scriptsize
  \Large
  \small
  \caption{Real cost comparison for confidential arithmetic and linear algebra operations at 112-bit security, $v_{100 \times 1}$ and $M_{100 \times 100}$.}
  \label{tab:arithmetic_cmp}
  \begin{tabular}{|c|c|c|c|c|c|c|} \hline
    &{AHE (Paillier)}&{SHE (RLWE)}&\multicolumn{2}{c|}{Garbled Circuits}&\multicolumn{2}{c|}{Secret Sharing}\\
    &Comp& Comp &Comp & Comm&Comp & Comm\\ \hline 
    Addition/Subtraction&0.01 ms & 0.2 ms & 37 ms &2 KB & 0.0 ms &0.0 KB\\ \hline
    Multiplication&0.05 ms & 39 ms & 138 ms & 40 KB & 1 s & 2 KB\\ \hline
    Comparison& 429 h & $10^5 h$ &37 ms & 2 KB & -&-\\ \hline
    Division & -  & - & 208 ms & 46 KB & - & -\\ \hline
    Vector Addition &0.6 ms & 0.2 ms & 36 ms &192 KB & 0.0 ms &0.0 KB\\ \hline
    Dot Product&6 ms & 39 ms & 5 s & 4 MB & 7 s & 195 KB\\ \hline
    Matrix-vector Multiplication& 1 s & 3 m & 8 m &396 MB& 7 s &290 KB\\ \hline
  \end{tabular}
\end{table} 

\begin{table} [h!] \centering
  \scriptsize
  \Large
  \small
  \caption{Examples for primitive switching strategies in hybrid composition of CML frameworks. }\label{tab:switch_cmp}
  \begin{tabular}{|c|c|p{4.0cm}|p{4.0cm}|} \hline
    \multirow{2}{*}{Framework}&\multirow{2}{*}{Primitive Switch}&\multirow{2}{*}{Operation Switch}&\multirow{2}{*}{Justification} \\
    & & & \\ \hline 
    Sharma and Chen \cite{sharma19} & SHE $\rightarrow$ GC& Matrix vector multiplication $\rightarrow$ Sign Check & Sign checking is impractically expensive with SHE whereas tolerable with GC. \\ \hline
    Nikolaenko et al. \cite{niko13sp} &AHE $\rightarrow$ GC &Matrix Additions $\rightarrow$ Cholesky's decomposition &  The operations of division and square root in Cholesky's decomposition were not feasible with the AHE scheme. \\ \hline
    Nikolaenko et al. \cite{niko13} & AHE $\rightarrow$ GC & Matrix Additions $\rightarrow$ Gradient Descent & Gradient descent involved multiplications, additions, and subtractions not entirely feasible with the AHE scheme. \\ \hline
    Mohassel et al. \cite{mohassel17}& SecSh $\rightarrow$ GC & Matrix-vector multiplication $\rightarrow$ Comparison & Comparison is impossible over randomly shared secrets leading the switch to the garbled circuits. \\ \hline
    Mohassel et al. \cite{mohassel17} & GC $\rightarrow$ SecSh & Comparison $\rightarrow$ Vector Subtraction & Use of garbled circuits for comparison was unavoidable however continuing GC on to vector subtraction would result in excessive cost overhead. \\ \hline
    Demmler et al. \cite{demmler15} & SecSh $\rightarrow$ AHE/OT& Data at rest $\rightarrow$ Multiplication & Multiplication with random shares required switching to either AHE or OT protocol involving the two parties in the frameworks. \\ \hline
    Riazi et al. \cite{riazi18} & SecSh $\rightarrow$ GC & Matrix matrix multiplication $\rightarrow$ ReLu computation & Sign checking is impossible over randomly shared secrets leading the switch to garbled circuits \\ \hline
    Riazi et al. \cite{riazi18} & GC $\rightarrow$SecSh & ReLu $\rightarrow$ Matrix vector multiplication & Use of garbled circuits for matrix vector multiplication is impractical. \\ \hline

  \end{tabular}
\end{table} 

\begin{table} [h!] \centering
  \scriptsize
  \Large
  \small
  \caption{Example CML methods that replace the expensive algorithmic components with their crypto-friendly versions.}\label{tab:simplifying_alogo}
  \begin{tabular}{|c|p{2.5cm}|p{2.5cm}|p{2.5cm}|p{3.7cm}|} \hline
    {Framework}&ML Algorithm&{Original Component}&{Crypto-friendly Component } & Benefits\\ \hline 
    Mohassel et al. \cite{mohassel17} & Logistic Regression, Neural Networks & Sigmoid, Softmax & ReLu & Avoids inversion and limits expensive confidential divisions to one. \\ \hline
    
    Graepel et al. \cite{graepel12} & LMC, Fisher's LDA & Divisions & Multiplications with incorporated division factors & Avoids division costs and simplifies the protocol. \\ \hline
    
    Nikolaenko et al. \cite{niko13sp} & Ridge Linear Regression & LU decomposition & Cholesky's decomposition & Reduces the cost complexity by half. \\ \hline
    Nikolaenko et al. \cite{niko13} & Matrix Factorization & Cholesky's Decomposition & Sorting based matrix factorization & Reduces the overall complexity from quadratic to within a polylogarithmic factor of the complexity in the plaintext \\ \hline
    Sharma and Chen \cite{sharma19} & Boosting & Decision Stumps & Random Linear Classifiers & Reduced number of comparisons and simplicity in learning. \\ \hline
    Naehrig et al. \cite{naehrig11} & Logistic Regression & Exponentiation & Taylor Expansion & Avoids costs involved in multiple levels of multiplications. \\ \hline    
    Sharma et al. \cite{sharma18tkde} & Spectral Clustering & Eigen decomposition & Eigen-approximation by Lanczos and Nystrom & Reduces complexity of the problem from $O(N^3)$ to $O(N^2)$. \\ \hline
  
      \end{tabular}
\end{table}

\end{backmatter}

\end{document}